\pdfoutput=1
\documentclass[11pt]{article}

\usepackage[]{emnlp2021}

\usepackage{times}
\usepackage{latexsym}

\usepackage[T1]{fontenc}
\usepackage[utf8]{inputenc}

\usepackage{microtype}

\usepackage{booktabs}
\usepackage{graphicx}
\usepackage{amsmath}
\usepackage{amsfonts}
\usepackage{float,stfloats}
\usepackage{enumitem}
\newcommand{\set}[1]{\left\{ #1 \right\}}

\title{Relating Neural Text Degeneration to Exposure Bias}

 \author{Ting-Rui Chiang \\ Carnegie Mellon University \\   \texttt{tingruic@andrew.cmu.edu}
         \And  Yun-Nung Chen \\ National Taiwan University\\  \texttt{y.v.chen@ieee.org}}
\begin{document}
\maketitle
\begin{abstract}
This work focuses on relating two mysteries in neural-based text generation: exposure bias, and text degeneration.
Despite the long time since exposure bias was mentioned and the numerous studies for its remedy,
to our knowledge, its impact on text generation has not yet been verified.
Text degeneration is a problem that the widely-used pre-trained language model GPT-2 was recently found to suffer from \cite{Holtzman2020The}.
% One salient problem is the repetitive loop that occurs pervasively when generating texts with likelihood-maximizing decoding strategies.
% There are some sophisticated hypotheses about the causation of this problem, but few of them have been well studied.
Motivated by the unknown causation of the text degeneration, in this paper we attempt to relate these two mysteries.
Specifically, we first qualitatively and quantitatively identify mistakes made before text degeneration occurs.
Then we investigate the significance of the mistakes by inspecting the hidden states in GPT-2.
% We show that the hidden state of the model tends to deviate to low density areas, implying that the mistakes are significant to GPT-2.
Our results show that text degeneration is likely to be partly caused by exposure bias.
We also study the self-reinforcing mechanism of text degeneration, explaining why the mistakes amplify.
In sum, our study provides a more concrete foundation for further investigation on exposure bias and text degeneration problems.
The scripts for our experiments are available at \url{https://github.com/MiuLab/Degenration-ExposureBias}.
\end{abstract}

\section{Introduction}
One mythology in neural text generation is \emph{exposure bias}~\cite{bengio2015scheduled,pomerleau1989alvinn,thrun1995learning}.
In the context of text generation, exposure bias refers to mistakes made by the model at the beginning of text generation, which may amplify, and lead the model to a state unseen in training time, and may thus cause misbehavior in the following generation.
Phenomena related to exposure bias were first observed in \cite{pomerleau1989alvinn} in the self-driving vehicles field.
After that, exposure bias was mainly discussed in the context of imitation learning \cite{thrun1995learning,ross2010efficient,ross2011reduction}.
In 2015, \citet{bengio2015scheduled} introduced it in the context of neural text generation.
However, its impact on text generation is questionable from both the empirical and the theoretical perspectives.
Empirically, despite the number of studies for its remedy \cite{bengio2015scheduled,huszar2015not,RanzatoCAZ15,lamb2016professor,yu2017seqgan,wiseman-rush-2016-sequence,schmidt-2019-generalization,zhang-etal-2019-bridging}, phenomena resulted from exposure bias have not yet been explicitly identified.
On the other hand, theories attained in the context of imitation learning may not be applicable to the above text generation tasks.
For example, \cite{ross2010efficient} shows a $O(T^2)$ trend of cost with respect to the number of steps $T$ in an episode.
It implies that the cost grows quadratically when $T$ is large.
However, most of natural language process tasks, e.g. machine translation and image captioning, do not generate very long text.
The impact of exposure bias is thus not clear for text generation tasks.

A younger mystery is the recently discussed enigma of \emph{text degeneration}~\cite{Holtzman2020The}.
It refers to the phenomenon in which bland or strange repetitive texts may be generated when the likelihood is the objective of generation, for example, when some commonly used strategies, such as greedy decoding and beam-search decoding, are used.
Especially, the prior work \cite{Holtzman2020The} observed such problems in GPT-2 \cite{radfordlanguage}, a pre-trained language model that has been shown useful in many NLP tasks \cite{Radford2018ImprovingLU,zhang2019dialogpt,petroni-etal-2019-language,talmor-etal-2019-commonsenseqa,see-etal-2019-massively}.
Despite many attempts proposed to address this issue \cite{Holtzman2020The,Welleck2020Neural,li2019don}, its root cause remains unknown.

\begin{table}[t]
\centering
    \small
    \begin{tabular}{p{.95\linewidth}}
    \toprule
    \textbf{GraphQL is an interesting technology originating at Facebook.} \emph{\textbf{It is a query language} that allows you to query a database and then query the database for the results.{\textbackslash}n {\textbackslash}n The query language is called QueryQL.} It is a query language  ... \\
    \midrule
    \textbf{We first saw Anki Overdrive, the company's follow-up to} the original game, in the early 2000s. It was a game that was a bit of a hit, and it was a game \textit{that was a bit of a hit} that was a bit of a hit that was a hit that ... \\
    \bottomrule
    \end{tabular}
    \caption{Randomly sampled examples generated by GPT-2 by greedy decoding. The \textbf{bold} part are the text conditioned on, and the \textit{italic} part are the text in the repetitive loop.}
    \label{tab:degeneration-sample}
    
\end{table}

Motivated by the unknown issues, we wonder whether text degeneration can be connected to the well-known exposure bias.
If text degeneration is the misbehavior caused by exposure bias, it actually provides us a perfect opportunity to identify the existence of exposure bias.
One of misbehavior of text degeneration is the occurrence of repetitive loops.
It is a phenomenon that a model tends to repeat a span of text during generation (an example is shown in Table~\ref{tab:degeneration-sample}.
This phenomenon is salient enough to be detected automatically, and occurs when greedy decoding strategy is used with high probability\footnote{93\% in our experiment.}.
The easiness of spotting can help the identification of exposure bias.
Therefore, this work aims at looking for the indications of exposure bias when repetitive loops are generated by the greedy decoding strategy.
We will focus on GPT-2, because it is the only publicly available language model trained on a massive amount of data at the time this work is done, and is widely used by the community.

To the best of our knowledge, this paper is the first work that attempts to relate text degeneration to exposure bias.
We first conclude two necessary conditions of its occurrence based on the intuition of exposure bias in literature in Section~\ref{sec:exposure-bias}.
We then inspect the two necessary conditions qualitatively and quantitatively.
In Section~\ref{sec:necessity1-exp}, we find that before text repeating starts, GPT-2 generates unnatural text.
In Section~\ref{sec:quantitative}, we show that the hidden states of GPT-2 deviate to an area less similar to the states generated by encoding real text.
The above observations satisfy the intuition of exposure bias that mistakes are made in the early stage and are amplified afterward.
According to the indications we discover, we conclude that exposure bias is likely to co-occur with repetitive loops.
Finally, we investigate how the mistakes are amplified after repetitive loops occur in Section~\ref{sec:mechanism}.
We discover the self-reinforcing mechanism of text degeneration.
The results provide a possible outline of how a model is trapped in repetitive loops.
These findings should be helpful for future studies on exposure bias and remedies for text degeneration.

\section{Related Work}

\subsection{Exposure Bias in Imitation Learning}
Imitation learning aims at imitating an expert policy $\pi^*$ by learning from trajectories generated by the expert, namely finding the policy
\begin{equation}
    \hat{\pi} = \arg \min_{\pi} \mathbb{E}_{s \sim d_{\pi^*}} I[\pi(s) = \pi^*(s)],
\end{equation}
where $d_{\pi^*}$ is the distribution of states visited by the expert policy $\pi^*$.
It is very similar to training a language model with maximum likelihood objective, and has succeeded in many applications \cite{pomerleau1989alvinn,schaal1999imitation,muller2006off,ratliff2006maximum}.
However, it was mentioned in \cite{pomerleau1989alvinn} that when a model makes a mistake and thus encounters a state that the expert rarely encounters, it may fail to recover from the mistake.
It was the first time the concept of exposure bias was mentioned.
Similar issues were also considered in \cite{thrun1995learning,daume2009search}.
\citeauthor{ross2010efficient} proved that the cost in a trajectory grows at the rate $O(T^2)$ instead of $O(T)$ if mistakes are made with a non-zero probability.
It can be seen as a theoretical analysis of exposure bias.
Nevertheless, in the context of text generation, the total number of steps in a trajectory is finite and is usually not large.
Therefore, it is still not clear how meaningful this growth rate of cost is for text generation tasks.
In \citet{ross2010efficient,ross2011reduction}, theoretically-grounded algorithms are proposed.
However, they require the access of expert policy to annotate the trajectories generated by the learnt agent.
It is generally not feasible in text generation tasks.

\subsection{Exposure Bias in Text Generation}
Then the concept of exposure bias is introduced in the context of text generation by \cite{bengio2015scheduled,RanzatoCAZ15}.
Since then, there have been many methods proposed to tackle this problem \cite{bengio2015scheduled,huszar2015not,RanzatoCAZ15,lamb2016professor,yu2017seqgan,wiseman-rush-2016-sequence,schmidt-2019-generalization,zhang-etal-2019-bridging,wang-sennrich-2020-exposure}.
They proposed their remedies based on the assumption that exposure bias is causing problems, and their approaches were justified by the improvement of performance when they are adopted.
However, to our knowledge, \citet{he2019quantifying} is the only study attempting to verify the impact of exposure bias, where they proposed metrics for estimating the impact of exposure bias in models.
Different from the prior work, this paper focuses on directly checking whether a specific phenomenon is the result of exposure bias.

\subsection{Neural Text Degeneration}
The term neural text degeneration was first defined recently in \citet{Holtzman2020The}, which focused on GPT-2.
Similar phenomenon was also observed in LSTM language models \cite{strobelt2018s}.
Regarding its causation, \citet{Welleck2020Neural} summarized three possible reasons about repetitive loops generated by GPT-2: i) The Transformer architecture of GPT-2 prefers repeating. ii) Repeating is an intrinsic property of human language. iii) The model is unable to model real language usage due to the fixed training corpora.
However, none of them have been proven theoretically or verified empirically.

Before this work, this phenomenon has not been linked to exposure bias, and thus remedies different from those for exposure bias are proposed.
\citet{Holtzman2020The} proposed sampling from the language model with nucleus sampling.
\citet{Welleck2020Neural} proposed to train neural language models with an unlikelihood as a regularization.
\citet{li2019don} further applied unlikelihood training on dialogue tasks.
Since in this work we discover the linkage between exposure bias and text degeneration, new approaches that specifically tackle exposure bias may be found effective for text degeneration in the future.

\section{Background and Notations}
To better elaborate the investigation of the above problems, background knowledge and notations are briefly introduced here.

\subsection{Real and Artificial Passages}
Considering that this paper focuses on analyzing the issues in text generation, we first define \emph{real} passages as natural language and \emph{artificial} passage as the generated language for following study.

\paragraph{Real Passages and Real Distribution}
\textit{Real passages} and \textit{real distribution} are related to training data.
Given $Y$ denoting the training set, a \textit{real passage} $y \in Y$ is a sequence of tokens $\set{y_1, y_2, \cdots, y_T}$, and \textit{real distribution} $P_Y$ is the distribution passages $y \in Y$ are drawn from, and it can be factorized as
\begin{equation}
    P_Y(y) = P_Y(y_1) \prod_{t=2}^T P_Y(y_t \mid y_1, y_2, \cdots, y_{t-1}).
\end{equation}

\paragraph{Artificial Passages and Artificial Distribution}
\label{sec:artificial-passages-dist}
% \textit{Generated passage} and \textit{generation generation distribution} are related to an auto-regressive (conditional) language model.
A \textit{artificial passage} $\hat{y}$ is a sequence of tokens $\set{\hat{y}_1, \hat{y}_2, \cdots, \hat{y}_T}$ generated by a model.
We denote the set of generated passages as $\hat{Y}$, where each $\hat{y}$ is generated based on the conditional probability, $P_{M}(\hat{y}_t \mid \hat{y}_1, \hat{y}_2, \cdots, \hat{y}_{t - 1} )$, predicted by an auto-regressive language model $M$ such as GPT-2.
We define \textit{artificial distribution} $P_{\hat{Y}}$ as the distribution of $\set{\hat{y} \in \hat{Y}}$ detailed below.
Note that $P_{\hat{Y}}$ could be different from $P_{M}$, depending on the decoding strategy used.
A decoding strategy is how a token $\hat{y}_t$ is chosen based on the conditional probability $P_{M}(\hat{y}_t \mid \hat{y}_1, \hat{y}_2, \cdots, \hat{y}_{t - 1} )$.
In this work, we considered the greedy strategy and the sampling-based strategies, including the top-k candidates at each step \cite{fan2018hierarchical}, nucleus sampling \cite{Holtzman2020The}. Details are included in the appendix.

\subsection{States of GPT-2}
\label{sec:gpt-2}
GPT-2 is a pre-trained language model constituted with $L$ layers of Transformer blocks \cite{vaswani2017attention}.
Considering that exposure bias is described as a general problem of neural text generation models, we pick GPT-2 as an example model for the study.
When the tokens $\set{y_t}_{t=1, 2, \cdots, T - 1}$, which we refer to as the \textit{conditioned passage}, are fed in, we denote the states outputted by each layers as
\begin{align}
    [&h^{(y)}_{1, 1}, h^{(y)}_{1, 2}, \cdots, h^{(y)}_{1, T}] = \nonumber  \\
    & \mathrm{transformer}_1(
        \mathrm{embedding}([y_1, \cdots, y_T])
    ),  \\
    [&h^{(y)}_{l, 1}, h^{(y)}_{l, 2}, \cdots, h^{(y)}_{l, T}] = \nonumber  \\
    & \mathrm{transformer}_l(
        [h^{(y)}_{l - 1, 1}, \cdots, h^{(y)}_{l - 1, T}]
    ) \nonumber \\
    & \forall l = 1, 2, ..., L.   
\end{align}
It predicts the conditional probability as
%$P_M(y_t \mid \set{y_t}_{t=1, 2, \cdots, T - 1})$ by
\begin{equation}
        P({y}_T \mid \{{y}_t\}_{t=1...T-1}) =
    \mathrm{softmax}( \mathrm{MLP}(h^{(\hat{y})}_{L, T - 1}).
\end{equation}
We refer to \textit{real states} as the states outputted when $y \sim Y$ is fed in,
and \textit{artificial states} as the states when $\hat{y} \sim \hat{Y}$ is fed in.
\textit{States of a token} $y_t$ refer to the set of states $\set{h^{(y)}_{l, t}}_{l=1, 2, \cdots, L}$.

\subsection{Repetitive Loops}
\label{sec:def-repetitive-loops}

Let the time step at which a passage $\hat{y}$ starts to repeat be $\rho$, and the length of the repeated part be $\lambda$.
Then a passage $\hat{y}$, where a repetitive loop occurs, is of the form
\begin{align}
    \hat{y} = & \hat{y}_1, \hat{y}_2, \cdots, \hat{y}_{\rho - 1}, \hat{y}_{\rho}, \cdots \hat{y}_{\rho + \lambda}, \nonumber \\
     & \hat{y}_{\rho}, \cdots \hat{y}_{\rho + \lambda}, \hat{y}_{\rho}, \cdots \hat{y}_{\rho + \lambda}, \cdots
\end{align}
We refer to the repeated part $\hat{y}_{\rho}, \cdots \hat{y}_{\rho + \lambda}$ as a \textit{looping sequence}.

\subsection{Exposure Bias}
\label{sec:exposure-bias}

In the literature, exposure bias was conceptually proposed \cite{bengio2015scheduled}, which is described as the discrepancy between the way the model is used during training and the way during inference.
When training, at the time step $t$, the model objective is to maximize the probability of the correct token $y_t$ conditioning on the \emph{real} past tokens $y_1, y_2, \cdots, y_{t-1}$.
However, during inference, $\hat{y}_t$ is predicted conditioning on the \emph{generated} past tokens $\hat{y}_1, \hat{y}_2, \cdots, \hat{y}_{t-1}$.
Therefore, mistakes in the early stage may lead the model to a state unseen in training time, and errors may consequently amplify quickly.

More explicitly, based on the description of bias in \citet{bengio2015scheduled}, we summarize the necessary conditions as follow: If some misbehavior, such as repetitive loop, starts at time $\rho$ is the result of exposure bias, then the two indications must be observed:
\begin{enumerate}
    \item \textbf{Mistakes are made in the early phase:}
    In the context of text generation, qualitatively, it means the unnatural sequence is generated before time step $\rho$.
    Quantitatively, it means that $P_Y(\hat{y}_1, \hat{y}_2, \cdots, \hat{y}_{\rho-1})$, the likelihood that the previous generated text is real, is low.
    \item \textbf{Mistakes are significant to the model:}
    The mistakes must be significant enough to lead the model to a state unseen in training time. Specifically, here we analyze the hidden states of GPT-2. We posit that, if some misbehavior is due to exposure bias, then the mistakes in the early stage should be significant enough to cause the model to generate an unseen state.
\end{enumerate}

\section{Relating Text Degeneration to Exposure Bias}

In this section, we investigate whether the conditions in Section~\ref{sec:exposure-bias} are satisfied when text degeneration occurs.

\subsection{Experimental Setting}
\label{sec:necessity1-exp}

As in \citet{Holtzman2020The}, we focus on the pre-trained language model GPT-2\footnote{We use the implementation from Hugging Face (\url{https://huggingface.co/transformers/index.html}).}.
GPT-2 is trained on the WebText dataset.
We use the training, validation and testing subsets of WebText released by OpenAI \footnote{\url{https://github.com/openai/gpt-2-output-dataset}}.
% We mainly focus on greedy decoding strategy in this section.

When generating passages, first 50 tokens from passages $y \in Y$ are given as the condition.
Therefore, for different conditions $y$, even if the decoding strategy is deterministic, the generated passages $\hat{y}$ could be different.
We empirically observe that repetitive loops tend to occur later when the number of conditioned tokens is greater.
We choose to condition on 50 tokens, so the sequences before repetitive loops are lengthy enough for analysis while the computation power required is affordable.

\subsection{Qualitative Inspection on Generated Tokens Prior to Repetitive Loops}
We inspect the first condition about exposure bias by \emph{subjectively} examining the passages generated before a repetitive loop occurs.
For each passage $\hat{y}$ generated by conditioning on $\set{y}_{t=1,2,\cdots,50}$, we compare the pair $\hat{y}_{t=51,\cdots,\rho - 1}$ (generated) and $y_{t=51,\cdots,\rho - 1}$ (real), where $\rho$ is the time step where the repeating sentence first appears.
We want to check if the model does make mistakes during $t=51,\cdots,\rho - 1$.
We manually examine 50 randomly sampled pairs
\footnote{We didn't use crowdsource, since this inspection needs to be done very carefully, and workers could be uncareful.}.
We observe that the generated passages are often less informative, less relevant or coherent to $\{y\}_{t=1}^{50}$.
As a result, without knowing which passage in the pair is real, we can still correctly identify the generated ones for 78\% of them.
We also inspect the sequence pairs from time $0$ to $\rho + \lambda - 1$, the time step \textit{after} which the model starts to repeat.
In that case, our correctness is even higher, up to 92\%.
Note that even though the annotation is not done by many people, the fact that the fake sentences can be identified accurately is suffice to claim that a portion of passages generated in the early stage are perceivably dissimilar to real language.
Namely $P_Y(\hat{y}_1, \hat{y}_2, \cdots, \hat{y}_{\rho-1})$ and $P_Y(\hat{y}_1, \hat{y}_2, \cdots, \hat{y}_{\rho + \lambda - 1})$ is low from human judgement.
Thus, qualitatively we can say mistakes are made before the repeating loop occurs.
It satisfies the first condition of expsure bias.
%And it is an indication that the first condition of exposure bias is satisfied.

\subsection{Quantitative Inspection on Generated Tokens Prior to Repetitive Loops}
\label{sec:quantitative}

We further inspect the first condition of exposure bias \emph{quantitavely} and \emph{objectively}.
We want to estimate $P_Y(\hat{y}_1, \hat{y}_2, \cdots, \hat{y}_{\rho-1})$, the likelihood $\hat{y}_1, \hat{y}_2, \cdots, \hat{y}_{\rho-1}$ is real.
However, the true $P_Y$ is not tractable.
Using an auto-regressive model to estimate the likelihood is not feasible either, since they may give higher probability to passages that is also generated by auto-regressive models and thus favor GPT-2.
Thus we use a pre-trained masked language model RoBERTa-Large \cite{liu2019roberta}.
It is trained non-autoregressively, so it does not favor auto-regressively generated passages.
Therefore it should be a good proxy estimating the realness of the passages generated by GPT-2.

Specifically, to estimate the likelihood of tokens in a passage, real passages and artificial passages with repetitive loops are fed in RoBERTa with 15\% randomly selected tokens masked.
Log likelihood of recovering the masked tokens is calculated.
To anneal the randomness due to the selection of masked tokens, this process is repeated 10 times for each passage.
Finally, the likelihood for each time step is averaged.

\begin{figure}
    \centering
    \includegraphics[width=\linewidth]{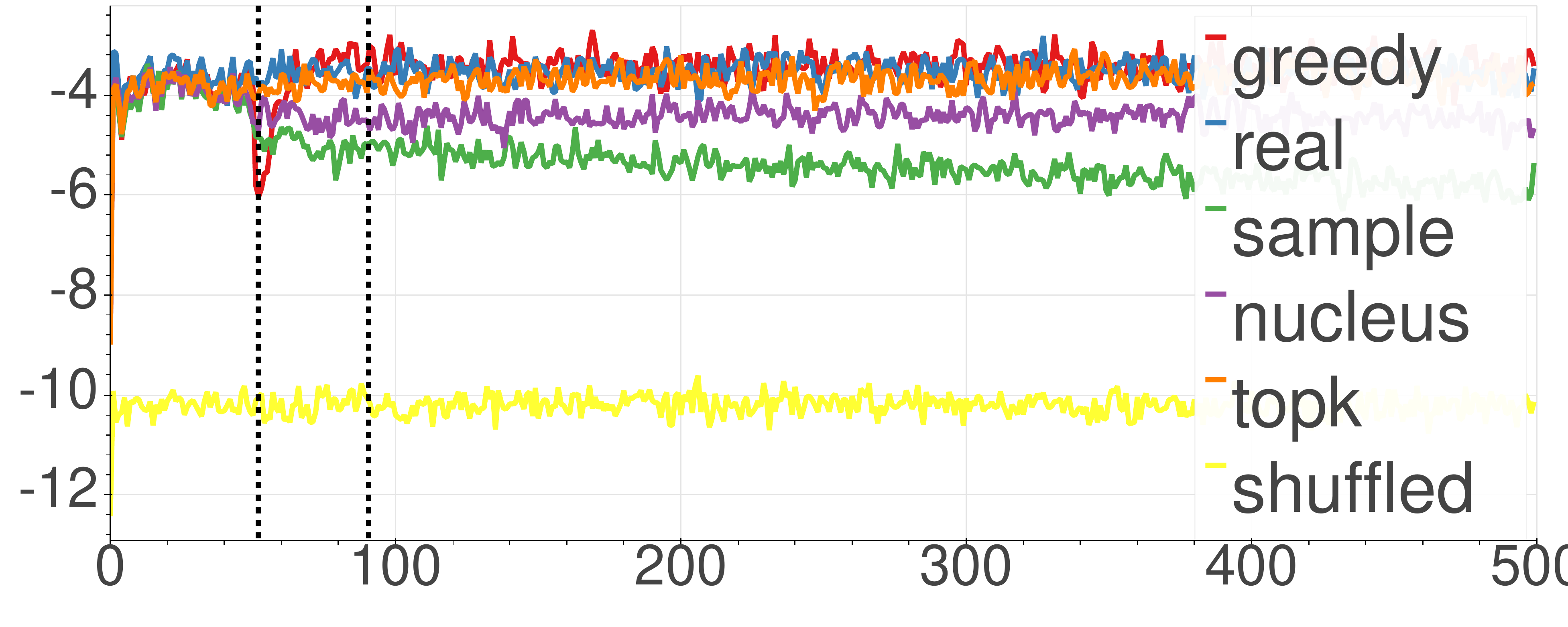}
    \caption{The average log likelihood (y-axis) predicted by RoBERTa at each time step (x-axis). The first dotted line is the averaged length of prefix, and the second one is the averaged $\rho$.} % The darker lines represents the average  log likelihood of tokens before repetitive loops occur, while the lighter lines represents the average of all tokens. The dashed vertical black line is the average length of the passages GPT-2 conditions on. }
    \label{fig:likelihood}
\end{figure}

Figure~\ref{fig:likelihood} shows that the likelihood of the generated passages is generally lower than real text starting from the time step where the conditioned passages end (dashed line).
Especially, even though the likelihood of the text generated with greedy decoding strategy grows after a few time steps, the likelihood drop significantly at the beginning.
Considering that the mask language model is sensitive to the context around the masked token, it may indicate that the text generated at the beginning is very unnatural.

\subsection{Significance of Mistakes Prior to Repetitive Loops}
\label{ref:significance}

We then check how significant the mistakes are to the GPT-2 model.
Though the previous sections have shown the existence of the mistakes in the early stage.
% However, if those mistakes do not cause the sequence to be significantly different from the real ones, then the following misbehavior is less likely to be caused by the mistakes, and thus exposure bias may be less likely to be a problem.
However, to cause misbehavior, the mistakes must be significant enough to cause GPT-2 to behave differently.
Therefore, we check how differently GPT-2 processes the generated text compared to the way it processes the real ones.

\subsubsection{Measuring the Significance of Mistakes}
To measure the significance of mistakes, we inspect the hidden states of GPT-2 when generating passages.
For each layer $l > 1$ and time step $t$, the artificial state $h^{(\hat{y})}_{l,t}$ is the result of applying the $\mathrm{transformer}$ function $l - 1$ times over the input sequence $\{ \hat{y}_{l - 1,\tau} \}_{\tau=1...t-1}$, which is the prefix of the artificial passage.
Therefore, if a artificial state $h^{(\hat{y})}_{l,t}$ is significantly dissimilar to any real states, then it implies that the generated passage $\{ \hat{y}_{l - 1,\tau} \}_{\tau=1...t-1}$ contains mistakes that are significant to the model, and that the mistakes do lead the model to an unseen state.
Thus, the similarity between the artificial states and the real state indicates how significant the mistakes in the passage are.

Specifically, we measure how many real state is similar to a artificial state.
It is done by counting the number of real states in the neighbor of the artificial state.
A lower number of real neighborhoods suggests that the artificial state is more unseen,
and thus implies higher significance of the mistakes.

Formally, given a hidden state $h_{l,t}^{(\hat{y})}$ at the time step $t$ in the layer $l$, we count the number of real states in a support set $H^Y_{l,t}$ which is close to $h_{l,t}^{(\hat{y})}$:
\begin{equation}
    N(h_{l,t}^{(\hat{y})}) = \left\vert \left\{ \left\lVert h_{l,t}^{(\hat{y})} - h \right\rVert_2 < r  \middle| h \in H^Y_{l,t} \right\} \right\rvert
    \label{eq:n-neighbor}
\end{equation}
where $r$ is the predefined radius.

\begin{figure}
    \centering
    \includegraphics[width=0.15\linewidth]{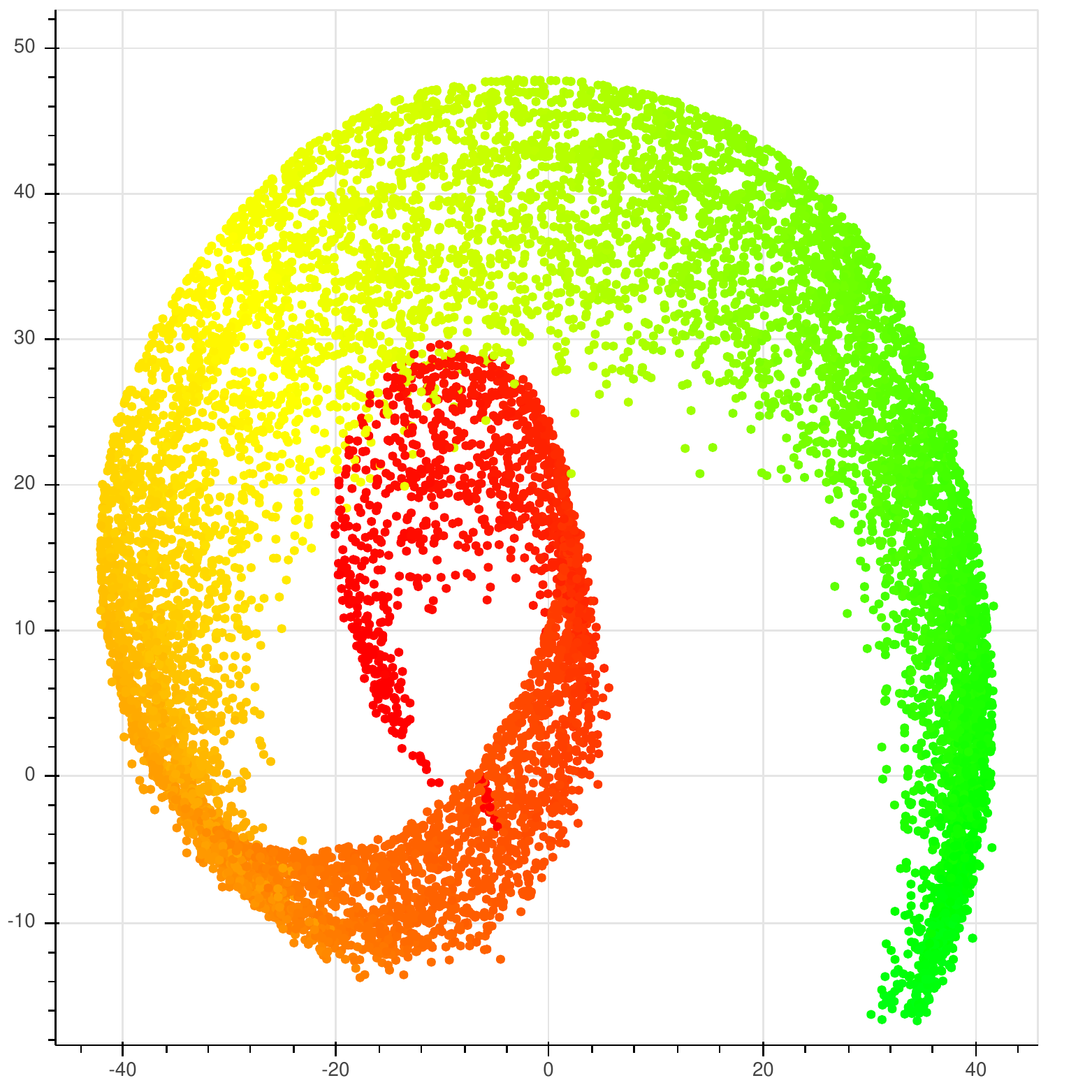}
    \includegraphics[width=0.15\linewidth]{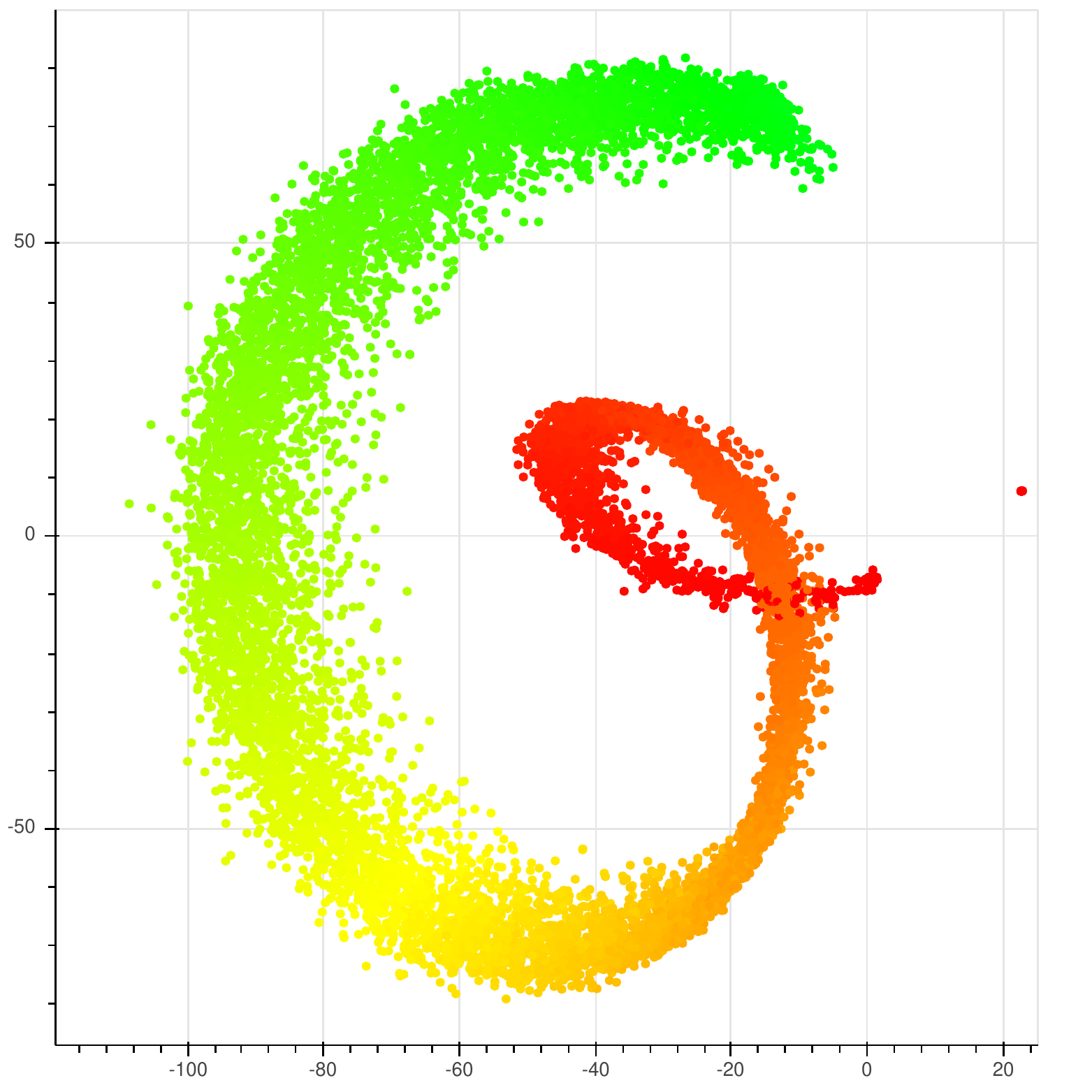}
    \includegraphics[width=0.15\linewidth]{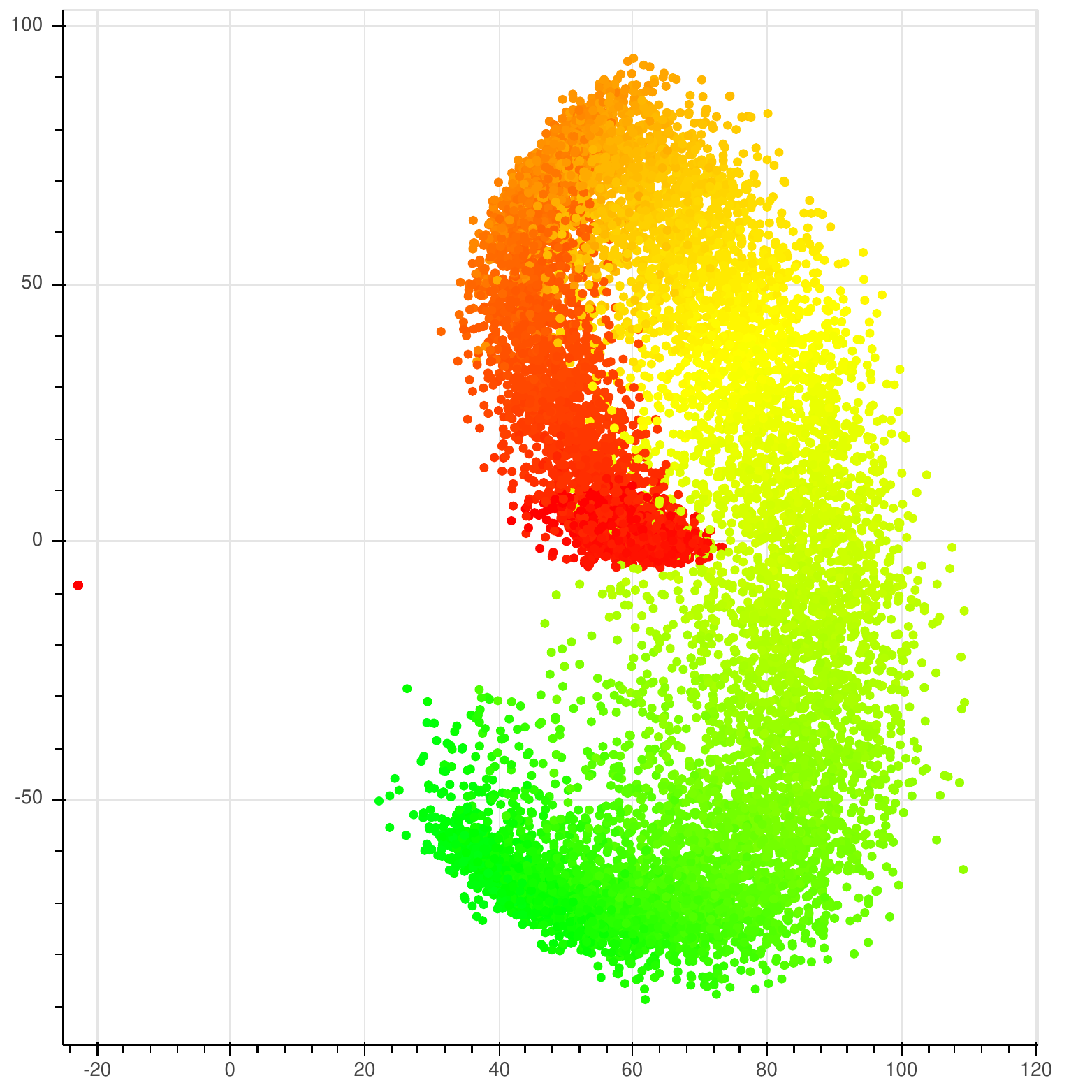}
    \includegraphics[width=0.15\linewidth]{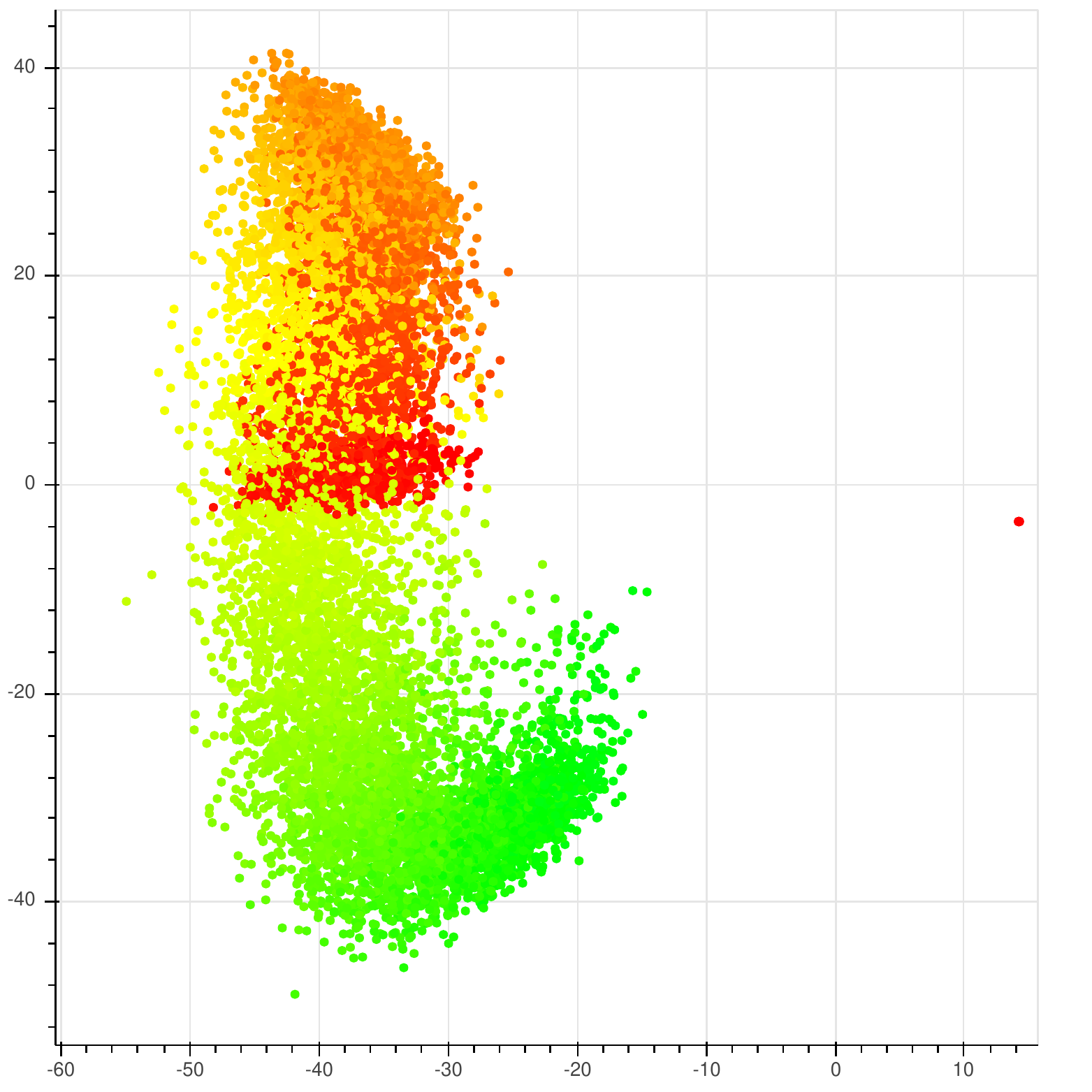}
    \includegraphics[width=0.15\linewidth]{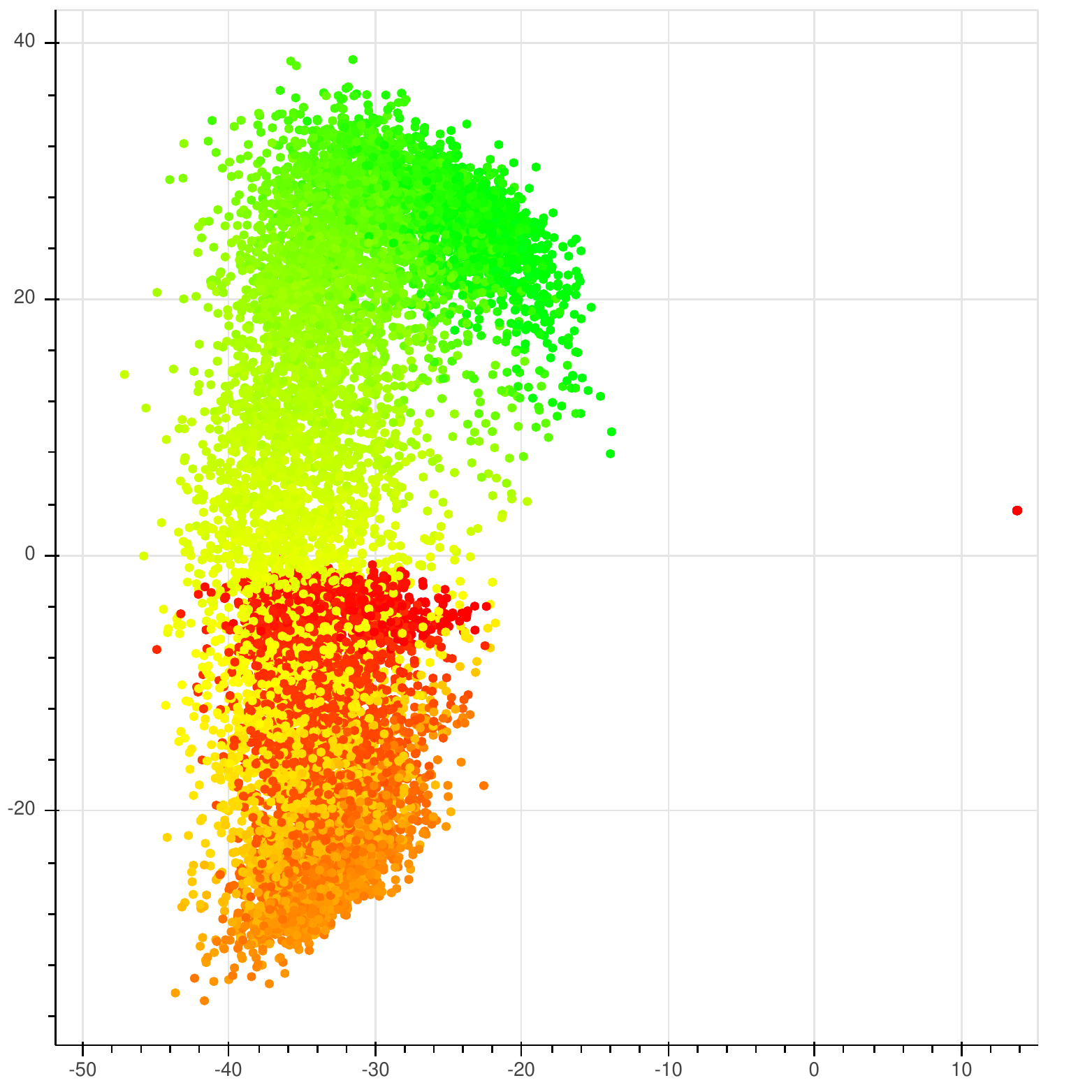}
    \includegraphics[width=0.15\linewidth]{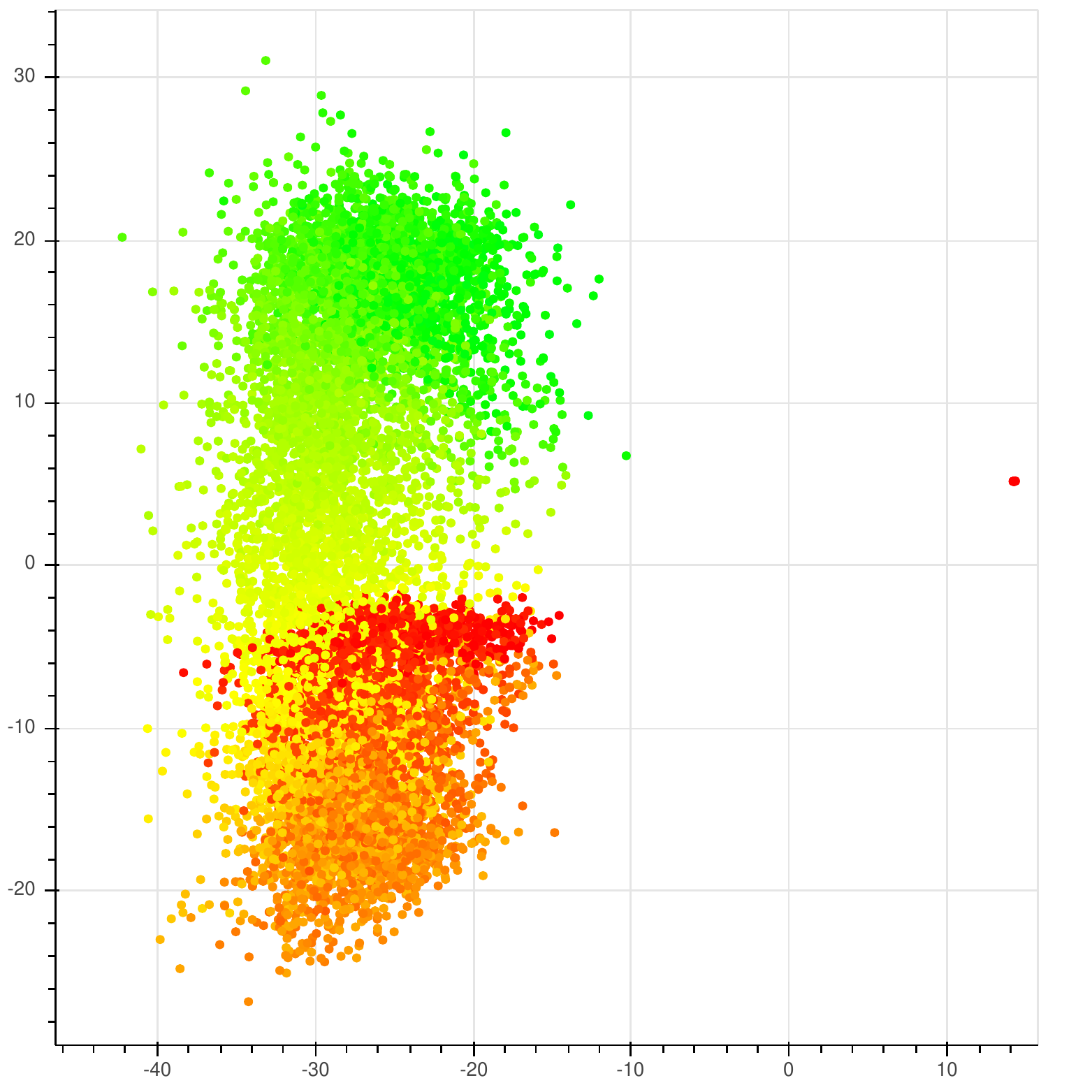}
    \caption{Hidden states projected to their first two principal components. Figures from left to right include states in layers 1, 3, 5, 7, 9, 11. Colors from red to green indicate the time steps from 0 to 512.}
    \label{fig:pca}
\end{figure}

We use different $H_{l,t}$ depending on the layer $l$ and the time step $t$ of the hidden state $h_{l,t}^{(\hat{y})}$ to be considered.
We compare $h_{l,t}^{(\hat{y})}$ only with the real states of the same layer, $H^Y_{l,t}$ only contains state of the same layer.
To reduces the required computation power, we also limit the set $H_{l,t}$ to the state of the tokens whose time step differ to $t$ by less than $\delta$\footnote{We use $\delta = 5$.}.
This limitation is reasonable, because we found the position of the states are time-step-dependent.
We found this by projecting the real states to their first two principle components with PCA \cite{pearson1901liii}.
As shown in figure~\ref{fig:pca}, states of nearby time steps are clustered together.
Formally, the support set of real neighbors is written as
\begin{equation}
    H^Y_{l,t} = \{ h_{l,\tau}^{(y)} \mid \tau \in [t - \delta, t + \delta], y \in Y \}.
    \label{eq:neighbor-set}
\end{equation}
Note that the constitution of $H^Y_{l,t}$ depends on a set of real passages $Y$. 
We will discuss the choice of $Y$ in the next section.

\subsubsection{Experimental Setting}

Roughly speaking, we perform our experiment with two sets of passages $Y_{sup}$ and $Y_{cond}$.
We use $Y_{cond}$ to generate real states, and from them we build $H^{Y_{cond}}_{l,t}$ for all layer $l$ and time step $t$.
$H^{Y_{cond}}_{l,t}$ can be used to evaluate any state of layer $l$ and $t$.
As for $Y_{cond}$, we use it to generate states to be evaluated.
By using the prefix of $y \in Y_{cond}$ as condition, we can use it to generate artificial passage $\hat{Y}$ and artificial states.
We also generate real states by encoding the whole $y \in Y_{cond}$ with GPT-2.
We expect that the states of $y \in Y_{cond}$ to be similar to the states of $y \in Y_{sup}$, while the artificial state $\hat{y} \in \hat{Y}$ to be dissimilar to the artificial ones.
Specifically, we conduct the experiment with the following steps:
\begin{enumerate}[nosep]
\item We prepare two disjoint sets of sequences $Y_{cond}$ and $Y_{sup}$.
There two sets are parts of the union of the training, validation and testing subsets of WikiText released by OpenAI (as described in Section~\ref{sec:necessity1-exp}).
\item For all $y$ in $Y_{sup}$, we collect a set of real states $h_{sup}$ by using GPT-2 to encode $y$.
These real states are used to construct the $H_{l,t}$ as mentioned in \ref{eq:neighbor-set}.
\item For all $y$ in $Y_{cond}$, we generate artificial sequences $\hat{y}$ by conditioning GPT-2 on the first 50 tokens of $y_{cond} \in Y_{cond}$.
We experiment with the generation strategies mentioned in Section~\ref{sec:artificial-passages-dist}.
The hidden states $\hat{h}$ are also collected.
\item For all $y$ in $Y_{cond}$, we also use GPT-2 to encode the whole passage $y$ and collect the states $h_{cond}$.
Since the sequences $y \in Y_{cond}$ are real, the states $h_{cond}$ are real too.
\item Finally, we evaluate how the states collected with $Y_{cond}$ are similar to the real states from $Y_{real}$.
We calculate $N(\hat{h})$, the numbers of artificial states' real neighbor in $h_{sup}$.
We also use $y_{cond}$ calculate $N(h_{cond})$.
% It provides us with a baseline number that we can compare $N(h_{cond})$ with.
It is referred to as "real" in Figure~\ref{fig:deviate-1024} and \ref{fig:deviate-shift}.
\end{enumerate}

We prepare $Y_{sup}$ and $Y_{cond}$ in two ways:
\paragraph{compare-seen}
The training split is used as $Y_{sup}$. It is \textit{seen} when training.
Real passages in the validation split and the testing split are used as $Y_{cond}$.
% In this setting, we measure how similar the artificial state is similar to the real states \textit{seen} in the training data.

\paragraph{compare-unseen}
The union of the validation split and the testing split is split into two disjoint subsets by ratio 9:1.
They are used as $Y_{sup}$ and $Y_{cond}$ respectively.
$Y_{sup}$ is \textit{unseen} when training.
% In this setting, we measure how similar the artificial state is similar to the states \textit{unseen} in the training data.

\subsubsection{Sanity Check}
We experiment with a set of \textit{shuffled} states as a sanity check of our approach.
It verifies whether the number of neighbors is an indicative measure of the significance of mistakes.
The \textit{shuffled} set is constructed by first shuffling the real passages in the $Y_{cond}$, and is then encoded with GPT-2.
The shuffled passages have the same 1-gram distribution as real natural language, but have low likelihood to be real.
We expect them to have low numbers of real neighbors.

The results show that the number of real neighbors is a good indication of mistakes for middle layers from layer 5 to layer 9 when $r=1024$ for both the compare-seen and compare-unseen settings.
For smaller $r \in {32, 64, 128, 256, 512}$, the results are not stable.
The average number of neighbors for different time steps at the seventh layer is plot in figure~\ref{fig:deviate-1024}.
We include the results of other layers in the appendix.
The figure shows that the number of neighbors of the \textit{shuffled} states are consistently low for all time steps.
It implies that the number of neighbors is indicative for detecting unreal passages.
However, it is less indicative when $R$ is small.
We posit that it is due to the high sparsity of the states due to their high dimensionality \footnote{Each state $\in \mathcal{R}^{1536}$}.

% Note that we do not consider the last layer since it is less indicative for the existence of exposure bias.
% Firstly, deviation in this layer will not be amplified since a state in this layer is not conditioned when generating other tokens.
% Secondly, states in the last layer can be similar to real states as long as they are similar to real token embeddings, since they are the ones fed into the linear layer for output probability.
% Thus even if the number of real neighbors in the last layer is high, the generated passage is still possible to be natural.

\subsubsection{Results}
\begin{figure}
    \centering
    \includegraphics[width=\linewidth]{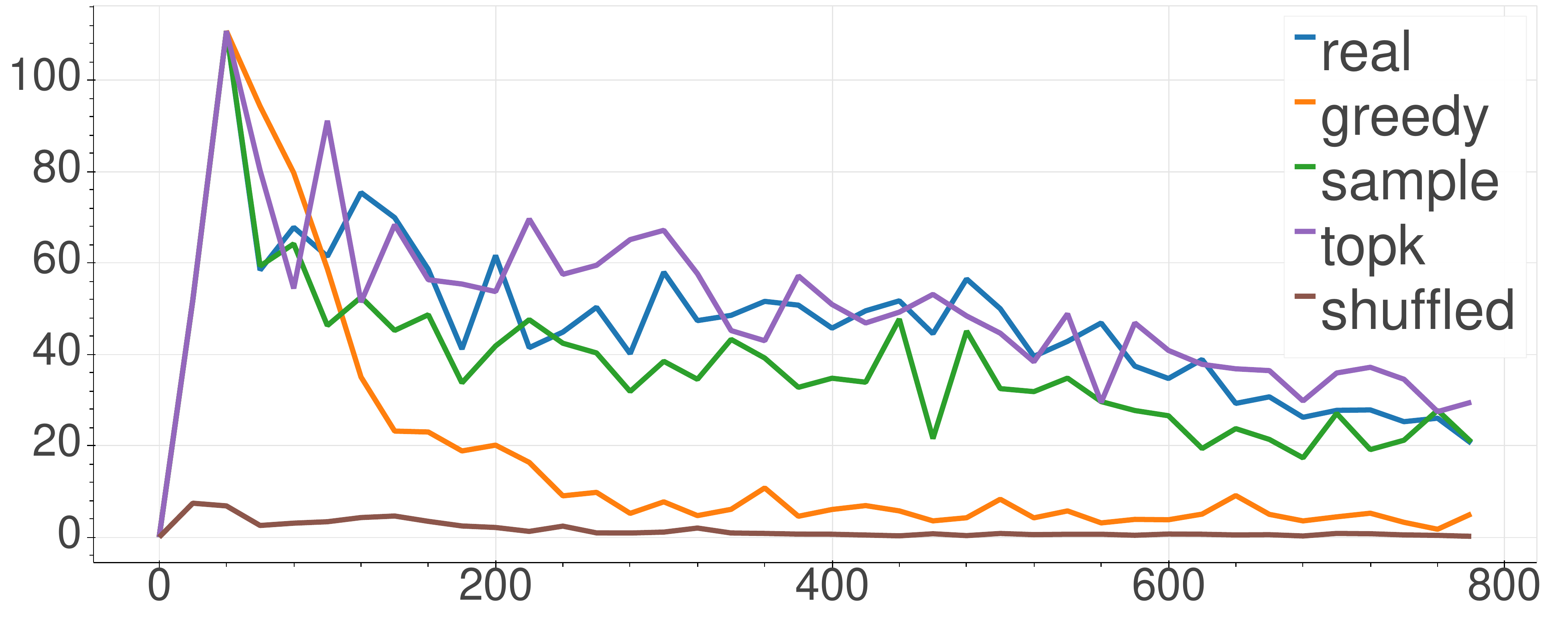}
    \includegraphics[width=\linewidth]{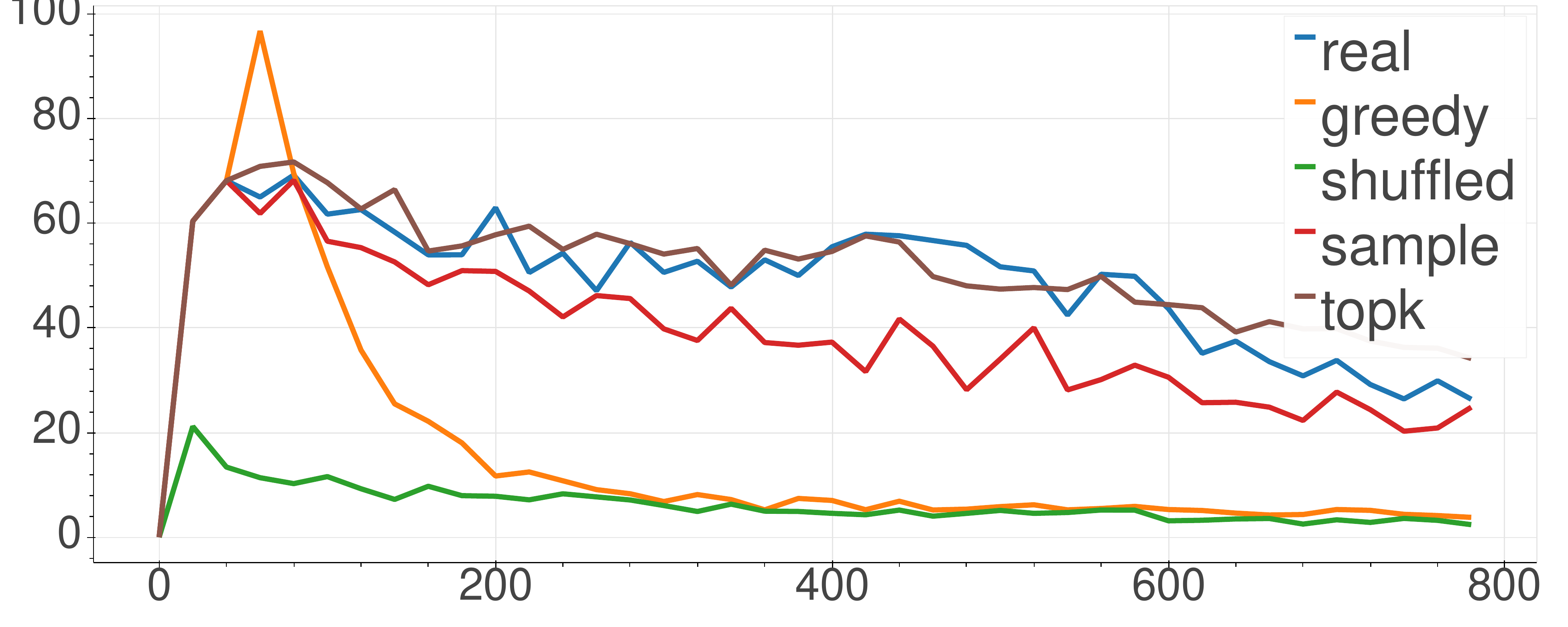}

    \caption{Number of neighbors in the \textit{seen-case} (top) and \textit{unseen-case} (bottom) at layer 7. The x-axis is the time step of the tokens. The y-axis is the number of real neighbors with the radius.}
    \label{fig:deviate-1024}
\end{figure}

Figure \ref{fig:deviate-1024} also plots the number of real neighbors ($h$) for states generated with greedy strategy and the sampling-based strategies ($\hat{h}$).
For the greedy strategy, the number of neighbors declines rapidly when the time step increases.
Note that we observe that repetitive loops occur in about 93\% of the sequences.
It shows that GPT-2 indeed fails to recover from mistakes, and the mistakes are amplified through time.
It is aligned with the description of exposure bias.
On the other hand, compared with real sequences (the control group), the number only decreases slightly when sampling-based strategies are used.
In contrast to the case of greedy decoding, repetitive loops are rarely observed when those sampling-based methods are used ($<$ 1\% for all of the strategies).
It implies that if GPT-2 has misbehavior when using those strategies, the misbehavior is less likely to be related to exposure bias.

\begin{figure}
    \centering
    \includegraphics[width=\linewidth]{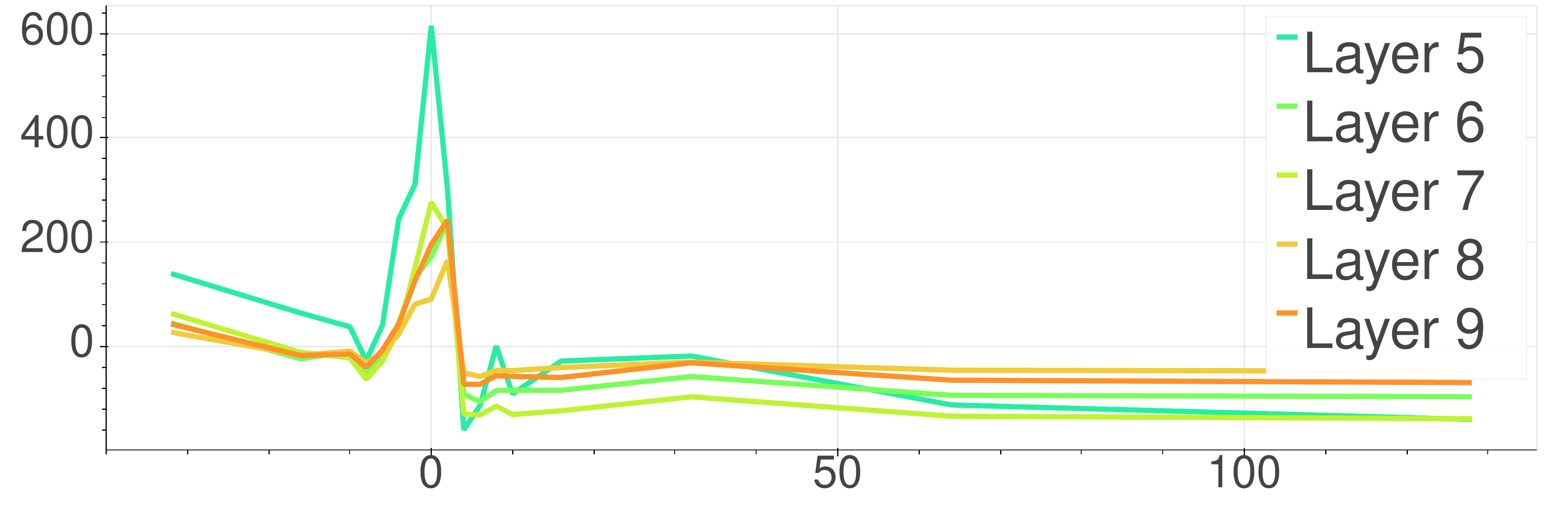}
    \includegraphics[width=\linewidth]{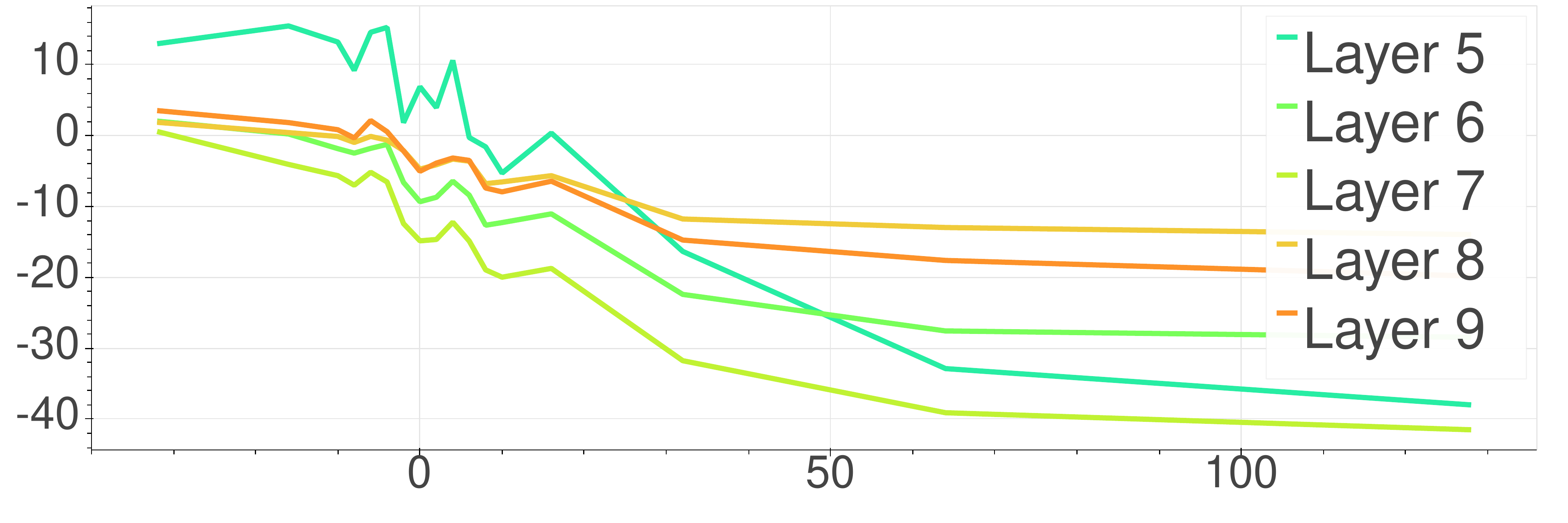}
    \caption{Number of neighbors for setting \textit{compare-seen} (upper) and \textit{compare-unseen} (lower). The x-axis is the time step of the tokens \emph{relative to} $\rho + \lambda$. The y-axis is $(n^{(\hat{y})}_{l, t} - n^{(y)}_{l, t})$.}
    \label{fig:deviate-shift}
\end{figure}

We further inspect the number of neighbors of the artificial state prior to the time step $\rho + \lambda$, when a repetitive loop starts.
We want to know whether the model does make significant mistakes before $\rho + \lambda$.
It is not shown in Figure~\ref{fig:deviate-1024}, as it only shows the significance of mistakes in the late stage.
To this end, we plot the number of neighbors again in Figure~\ref{fig:deviate-shift}.
Different from Figure~\ref{fig:deviate-1024}, in Figure~\ref{fig:deviate-shift}, the $x$-axis is the time step \emph{relative to $\rho + \lambda$}, so the significance of mistakes before repetitive loops can be manifested.
In particular, we compare the number of real neighbors around the real states and the artificial states.
Formally, for each artificial passage $\hat{y}$ conditioning on $y_{1, 2, \cdots, 50}$, we compare the number of neighbors around the state of $n^{(\hat{y})}_{l, t}$, and the state of the real passage following the condition $n^{(y)}_{l, t}$.
Here we set the y-axis of Figure~\ref{fig:deviate-shift} to be the difference $(n^{(\hat{y})}_{l, t} - n^{(y)}_{l, t})$.

Surprisingly, in Figure~\ref{fig:deviate-shift}, the \textit{compare-seen} and \textit{compare-unseen} settings show different trends.
At the beginning, the number of neighbors decreases relatively slowly in both of the two settings.
At around $x = -10$, the number in both of them drop to less than zero.
It indicates that at this time step, some significant mistakes are made.
However, the number in the \textit{compare-seen} setting dramatically grows while the number continues decreasing in the \textit{compare-unseen} setting.
The low number of neighbors in the \textit{compare-unseen} indicates the low realness of the generated passages.
The high number of neighbors in the \textit{seen-setting} indicates that the model encodes those unreal passages to space close to the states of training data.
It may imply that, at this moment, the model fails to generalize, so it incorrectly encodes the unreal passages as seen ones.
Finally, the mistakes are amplified.
Consequently, the number in both of the settings drops to less than zero.
In sum, Figure~\ref{fig:deviate-shift} shows the significance of mistakes made before it starts repeating a looping sequence.
Therefore, the second indication of exposure bias is observed.

\begin{table}[h!]
    \centering
    \caption{Similarity between the conditioned passage and the generated passage of the same length.}
    \label{tab:similarity-condition}
    \begin{tabular}{l c}
        \toprule
        \bf Conditioned Sentences & \bf Similarity (mean/std) \\
        \midrule
        Looping sequences & 0.7327 / 0.3226 \\
        First sentences & 0.2157 / 0.1911 \\
        Last sentences & 0.1837 / 0.1848 \\
        \bottomrule
    \end{tabular}
\end{table}

\begin{table}[ht!]
    \centering
    \small
    \begin{tabular}{cccc}
    \toprule
        \bf Repeat \# & \bf Looping Seq. & \bf First Sent. & \bf Last Sent. \\
    \midrule
        1 & 0.451 / 0.368 & 0.148 / 0.155 & 0.131 / 0.144 \\
        2 & 0.681 / 0.423 & 0.331 / 0.373 & 0.337 / 0.377 \\
        3 & 0.888 / 0.282 & 0.492 / 0.435 & 0.578 / 0.431 \\
    \bottomrule
    \end{tabular}
    \caption{The ROUGE-L (mean/std) between the sentences in the generated repetitive loops and $x$, when GPT-2 conditions on the pattern $c, x, \cdots, x$. }
    \label{tab:similarity-condition-n}
\end{table}

\section{Mechanisms after a Repetitive Loop Starts}
\label{sec:mechanism}

While the above experiments show the indications of exposure bias, in this section we further investigate how the early stage mistakes cause the model to degenerate.
Figure~\ref{fig:deviate-shift} indicates some mistakes are made prior to time step $\rho + \lambda$.
Thus, in this section, we investigate the characteristics of the sequence generated prior to $\rho + \lambda$, the looping sequence $\hat{y}_{\rho} \cdots \hat{y}_{\rho + \lambda}$ (as defined in Section~\ref{sec:def-repetitive-loops}).

\subsection{The Looping Sequence is Loop-Inducing}
\label{sec:loop-induce}
%$\hat{y}_{\rho}, \cdots \hat{y}_{\rho + \lambda}$ is Loop-Inducing}
We investigate how the looping sequences are loop-inducing by using them as conditions when generating text.
We construct a \textit{looping sequence} set that is constituted with all looping sequences generated when conditioning on the first 50 tokens of real sequences.
In a generated sequence $\hat{y}$, since $\hat{y}_{\rho}$ may not be a start point of a grammatical sentence, we use the sequence $\hat{y}_{\rho + \delta + 1}, \cdots, \hat{y}_{\rho + \lambda} \hat{y}_{\rho} \cdots \hat{y}_{\rho + \delta}$, where $\delta$  is chosen based on the punctuation in it \footnote{For example, if the looping sequence is "an apple. It is", we use "It is an apple."}.
As control groups, we also construct two real sequence sets, \textit{first sentence} set and \textit{last sentence} set.
They consist of the first sentence and the last sentence of the articles in WikiText validation split and testing split.

To measure how those sequences are looping-inducing, we calculate the similarity between $x$ and $\hat{y}$, where $x$ is the sequence used as condition, and $\hat{y}$ is the generated passage.
Specifically, we measure ROUGE-L~\cite{lin-2004-rouge}\footnote{We use the implementation in \url{https://github.com/google-research/google-research/tree/master/rouge}.} between $x$ and the first $\mathrm{length}(x)$ tokens of $\hat{y}$.
A higher score implies higher similarity, and thus more looping-inducing.
Results shown in Table~\ref{tab:similarity-condition} indicate that looping sequences are indeed more loop-inducing.

\subsection{Any Repeating Sequence is Loop-Inducing}
\label{sec:repeat-inducing}

We further discover that any sequence that is repeated is loop-inducing, regardless of contexts.
We create the conditioned sequence by concatenating $c$ with $x$ repeated from 1 to 3 times, where
$c$ is the first 5 sentences from a random article of WebText, and $x$ is either from the \textit{looping sequence} set or the real sets.
Measurement, the same as in Section~\ref{sec:loop-induce} is applied on $x$ and the generated passages.
The results are shown in Table~\ref{tab:similarity-condition-n}, and it shows that even when the conditioned sequence is real, it is more loop-inducing if it is repeated more times.

\subsection{The Self-Reinforcing Mechanism of Text Degeneration}
In sum, in this section, we discover the self-reinforcing mechanism of text degeneration.
First, Section~\ref{sec:loop-induce} a looping sequence is loop-inducing.
Thus, after a looping sequence is generated, it is likely to be repeated.
Second, Section~\ref{sec:repeat-inducing} shows that when a sequence is repeated, then GPT-2 would be more likely to continue repeating it.
Therefore, it shows how GPT-2 fails to recover from the mistake.

\section{Conclusion}
In conclusion, we provide a deeper insight into the relation between exposure bias and text degeneration.
We qualitatively and quantitatively show that mistakes are indeed made in the early stage of generation.
In Particular, some significant mistakes are made prior to $\rho + \lambda$, the time step when the model starts repeating.
We then show why the model fails to recover from the mistakes.
The looping sequence, which is the sequence generated prior to $\rho + \lambda$, and repeated sequences are looping-inducing.
That is how the model fails to recover from the mistakes, and how the mistakes amplify.
% Therefore, the looping phenomenon of text degeneration is likely to be a result of exposure bias.

Our contributions are four-fold: 1) We explicitly formulate the necessary indications for the detection of exposure bias.
2) For each condition, we design the associated experiments for validation.
3) By the experiments, we show that text degeneration is likely to be partly caused by exposure bias.
4) Finally, we provide a possible explanation how GPT-2 fails to recover from the mistake.
Our formulation and the conducted experiments build a solid foundation for future study on exposure bias.

\section*{Acknowledgements}
We would like to than Ting-Yun Chang for in-depth discussions. We are thankful to the anonymous reviewers for their insightful comments on the paper.
This work was financially supported from the Young Scholar Fellowship Program by Ministry of Science and Technology (MOST) in Taiwan, under Grant 110-2636-E-002-003.

\bibliographystyle{acl_natbib}
\bibliography{anthology,emnlp2020}

\appendix
\appendix

\section{Sample-based Decoding Strategies}
\begin{itemize}
    \item Sampling: $\hat{y}_t$ is directly sampled from the conditional probability $P_{M}(\hat{y}_t \mid \hat{y}_1, \hat{y}_2, \cdots, \hat{y}_{t - 1})$.
    \item Top-$k$ sampling \cite{fan2018hierarchical}: At the time step $t$, $\hat{y}_t$ is sampled from the conditional probability: % $\tilde{P}_{M}(\hat{y}_t | \{ \hat{y}_T \}_{t=1...T-1})$, where
    \begin{align}
        &{P}_{Y} (\hat{y}_t \mid \hat{y}_1, \hat{y}_2, \cdots, \hat{y}_{t - 1} ) \propto \nonumber \\
        &\begin{cases}
            P_{M}(\hat{y}_t \mid \hat{y}_1, \hat{y}_2, \cdots, \hat{y}_{t - 1}) & \text{if $\hat{y}_t \in$ top-$k$,} \\ %of $P_{M}(\cdot \mid \{ \hat{y}_T \}_{t=1...T-1})$} \\
            0 & \text{otherwise.}
        \end{cases}
    \end{align}
    \item Nucleus sampling \cite{Holtzman2020The}: At the time step $t$, $\hat{y}_t$ is sampled from the conditional probability % $\tilde{P}_{M}(\hat{y}_t | \{ \hat{y}_T \}_{t=1...T-1})$, where
    \begin{align}
        &{P}_{Y}(\hat{y}_t \mid \hat{y}_1, \hat{y}_2, \cdots, \hat{y}_{t - 1})
        \propto \nonumber \\
        &\begin{cases}
            P_{M}(\hat{y}_t \mid \hat{y}_1, \hat{y}_2, \cdots, \hat{y}_{t - 1} ) & \text{if $\hat{y}_t \in V^{(p)}$} \\
            0 & \text{otherwise.}
        \end{cases},
    \end{align}
    and for a predefined $p \in (0, 1]$, $V^{(p)}$ is the minimal set that satisfies
    \begin{equation}
        \sum_{v \in V^{(p)}} P_{M}(v \mid \hat{y}_1, \hat{y}_2, \cdots, \hat{y}_{t - 1}) \geq p
    \end{equation}
\end{itemize}

\section{Dataset}

We use the subsets of WebText released by OpenAI (\url{https://github.com/openai/gpt-2-output-dataset}).
It is an English dataset.
There are 25000, 5000, 5000 passages in the train, validation, testing splits respectively.
For experiments in Section~\ref{sec:quantitative} and Section~\ref{ref:significance}), we only use the passages with more than 512 tokens.
After passages with less than 512 tokens are removed, there are 5269 passages in the union of the validation split and the testing split.

\section{Detail of Experiments}

\subsection{Quantitative Inspection of Generated Tokens Priors to Repetitive Loops (Section~\ref{sec:quantitative})}
Implementation of RoBERTa from Python package transformers 2.8.0 by Hugging Face (\url{https://huggingface.co/transformers/}) is used.

\subsection{Significance of Mistakes Prior to Repetitive Loops (Section~\ref{ref:significance})}
We use Faiss~\cite{JDH17} to calculate the number of neighbor vectors within a radius.
For Figure~\ref{fig:deviate-1024}, the number of neighbors is calculated for 20 time steps.
For Figure~\ref{fig:deviate-shift}, the number of neighbors is calculated at time steps \{-32, -16, -10, -8, -6, -4, -2, 0, 2, 4, 6, 8, 10, 16, 32, 64, 128\} relative to $\rho + \lambda$.

\paragraph{Seen-setting:}
We sampled 2500 passages from the WebText training split.
Each line in Figure~\ref{fig:deviate-shift} is the average over 500 passages generated by each decoding strategy.
The result in Figure~\ref{fig:deviate-shift} is the average over 1000 passages.

\paragraph{Unseen-setting:}
We first combine the validation split and the testing split as the set of all real unseen text $\bar{Y}$.
Then we split it into 10 equal-sized subsets $\bar{Y}_1, \bar{Y}_2, \cdots, \bar{Y}_{10}$.
We repeat the following process 3 times:
\begin{itemize}
    \item From $\{\bar{Y}_1, \bar{Y}_2, \cdots, \bar{Y}_{10}\}$, a subset $Y_{real}$ is selected, and the rest $\bar{Y} \textbackslash Y_{real}$ is used as the support set $Y_{support}$.
    \item Real states are collected by encoding passages in $Y_{support}$ with GPT-2. When we are calculating the number of neighbors, only these real states are counted.
    \item Artificial passages are generated by conditioning on the first 50 tokens for passages in $Y_{real}$ using the decoding strategies.
    \item The number of neighbors is calculated for each decoding strategies.
\end{itemize}
Finally, the result is averaged to plot Figure~\ref{fig:deviate-1024} and Figure~\ref{fig:deviate-shift}.

\subsection{Automatic Detection of Looping Sequence}
Given a passage $x_1, x_2, \cdots, x_T$, we first search for the length of a repetitive loop by comparing $x_{T - \lambda + 1}, \cdots x_T$ and $x_{T - 2\lambda + 1}, \cdots x_{T - \lambda}$ for $\lambda = 4, 5, \cdots, l / 2, 1, 2, 3$.
If there exists some $\lambda$ such that $x_{T - \lambda + 1}, \cdots, x_T = x_{T - 2\lambda + 1}, \cdots, x_{T - \lambda}$, then we search $\rho$ as the first place such that $x_{\rho + i \lambda}, \cdots, x_{\rho + (i + 1)\lambda - 1} =  x_{T - 2\lambda + 1 + \delta}, \cdots, x_{T - \lambda + \delta}$ for some $\delta \in [0, \lambda - 1]$ and all $i$ such that $\rho + i \lambda < T$.

\section{Computing Infrastructure}
Each of our experiments were run on a workstation with 187 GiB RAM.
A workstation is equipped with either two Intel Xeon 5218 CPUs or two Intel Xeon 4110 CPUs.
Every experiment can be run with 1 Nvidia GTX 2080Ti GPU.

\begin{figure*}
    \centering
    \includegraphics[width=0.49\linewidth]{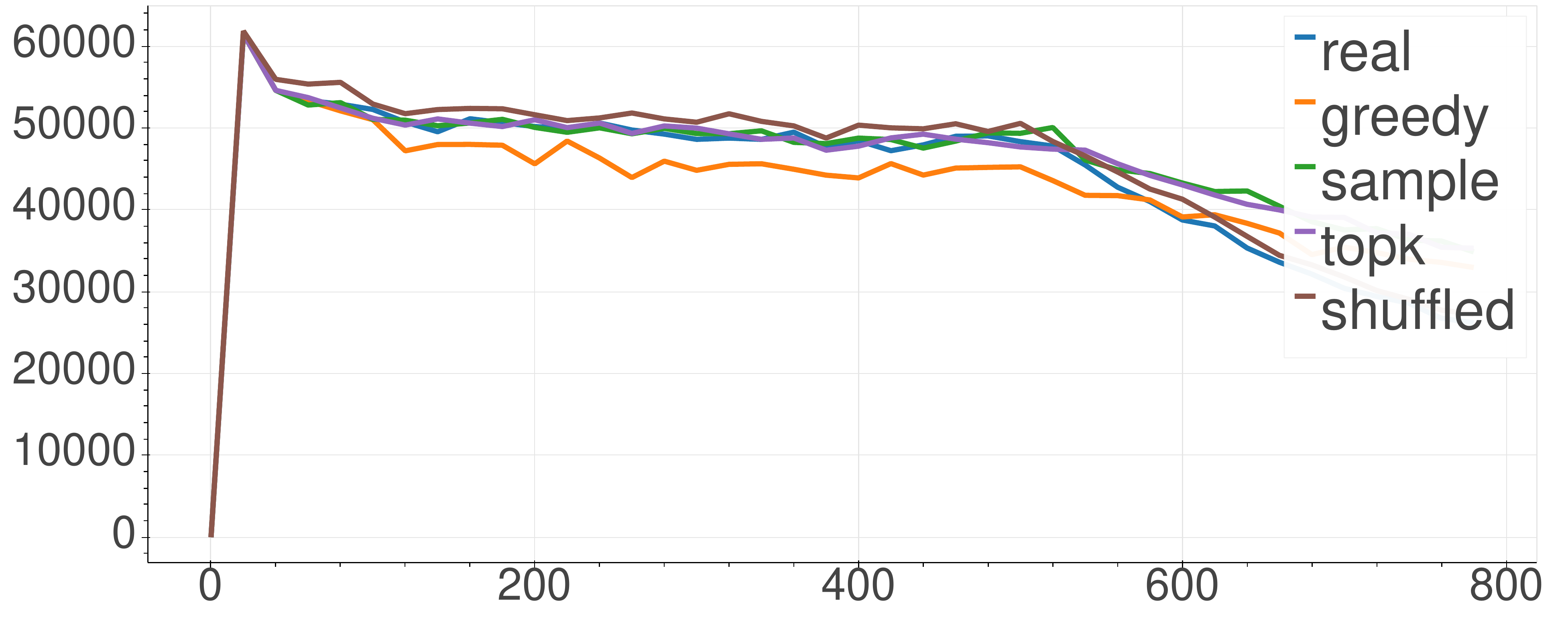}
    \includegraphics[width=0.49\linewidth]{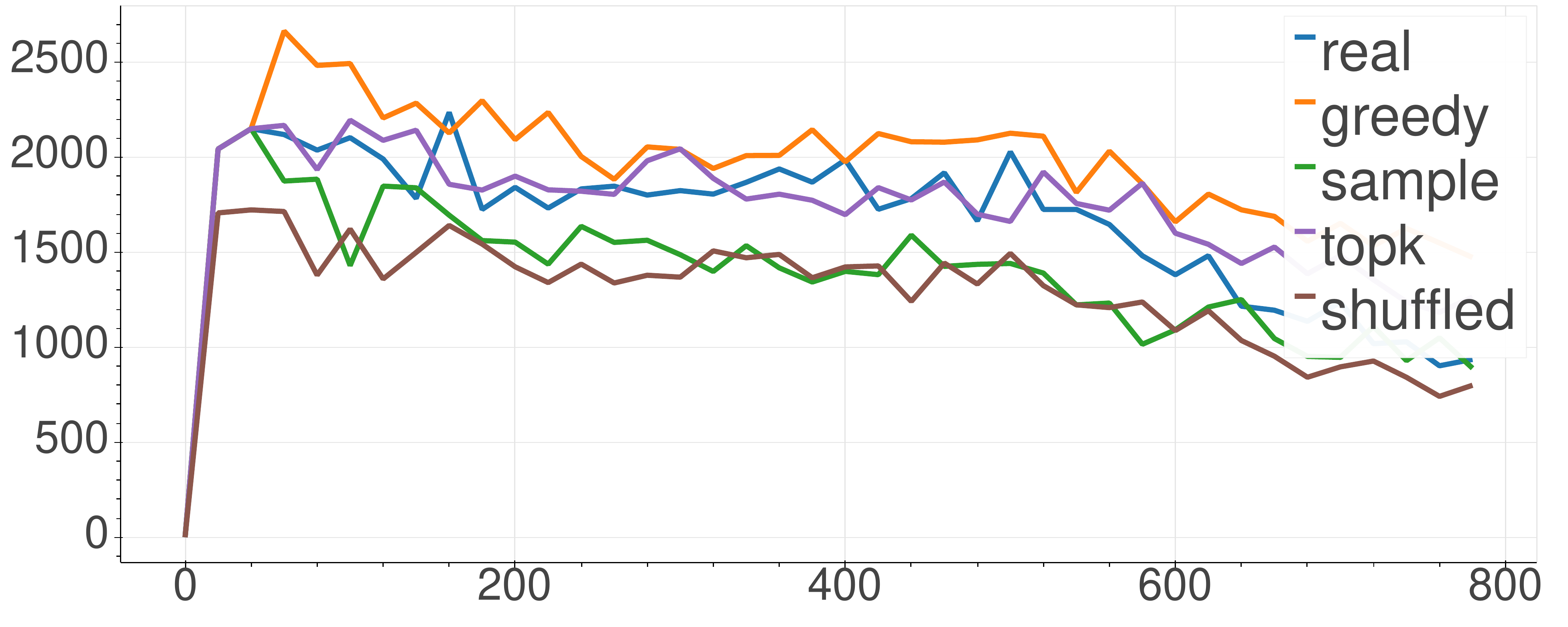}
    \includegraphics[width=0.49\linewidth]{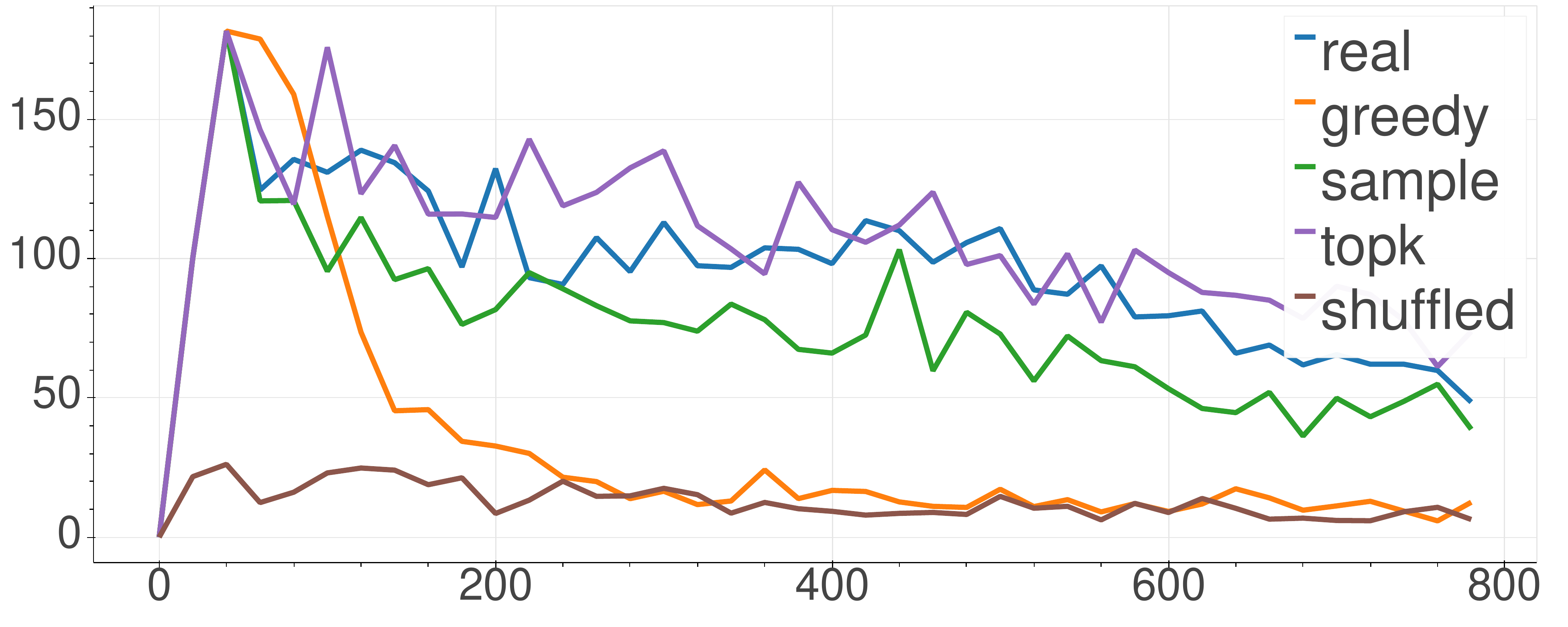}
    \includegraphics[width=0.49\linewidth]{figures/n-neighbor-1024-layer-7.pdf}
    \includegraphics[width=0.49\linewidth]{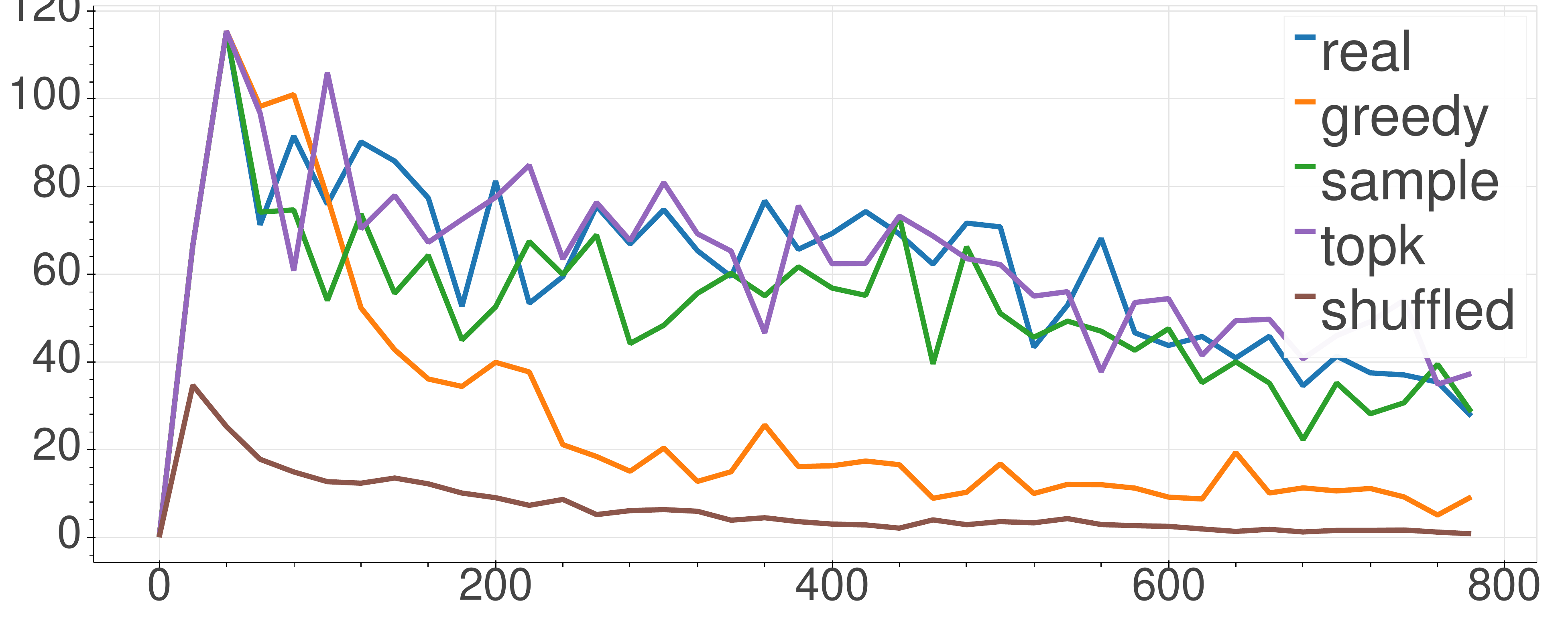}
    \includegraphics[width=0.49\linewidth]{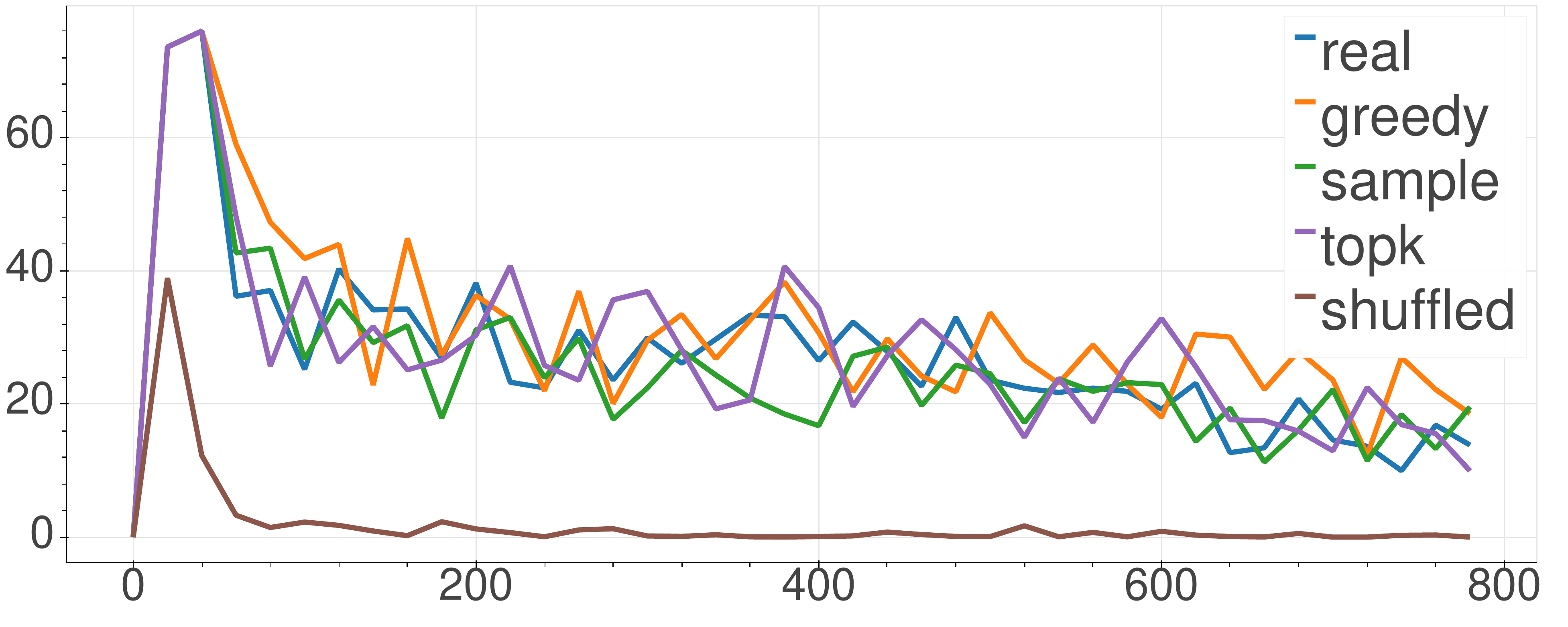}
    \caption{Number of neighbors for \textit{compare-seen} setting. The figures are the number of layer 1, 3, 5, 7, 9, 11, from left to right, top to bottom. The x-axis is the time step of the tokens. The y-axis is the number of real neighbors with the radius.}
\end{figure*}

\begin{figure*}
    \centering
    \includegraphics[width=0.49\linewidth]{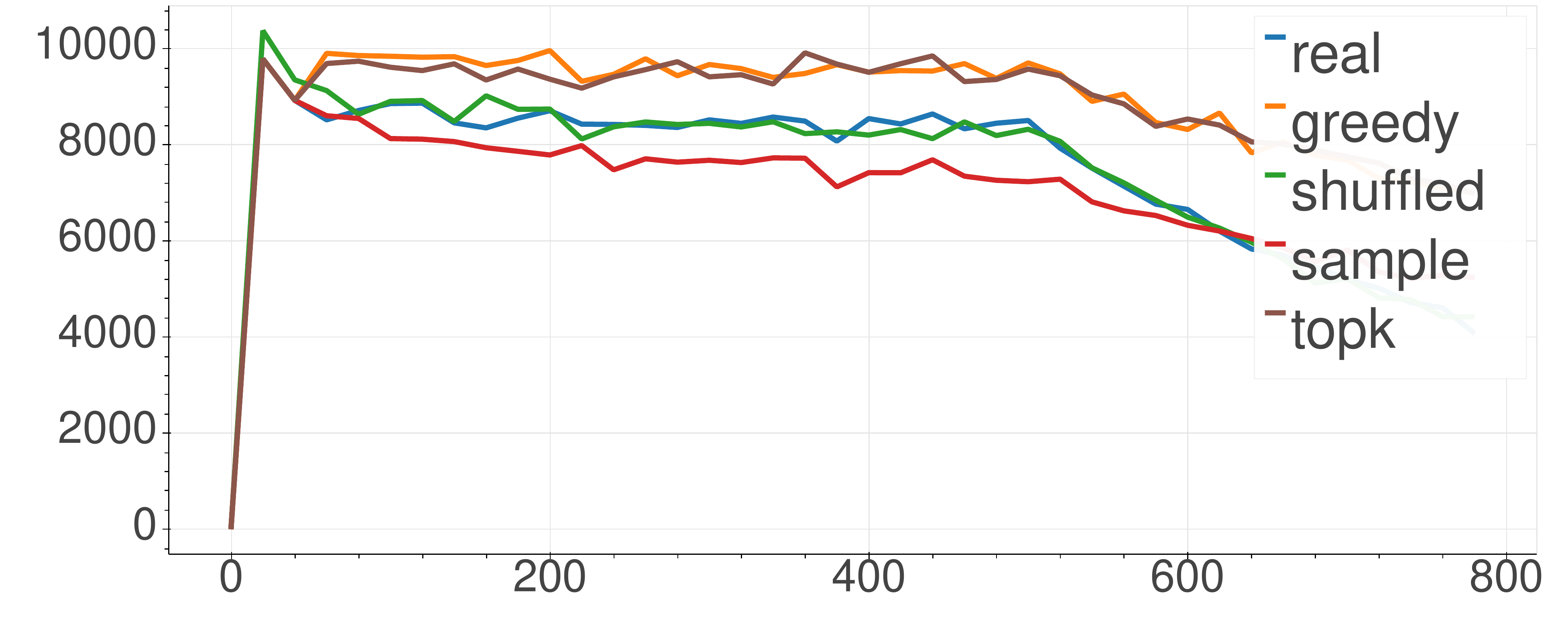}
    \includegraphics[width=0.49\linewidth]{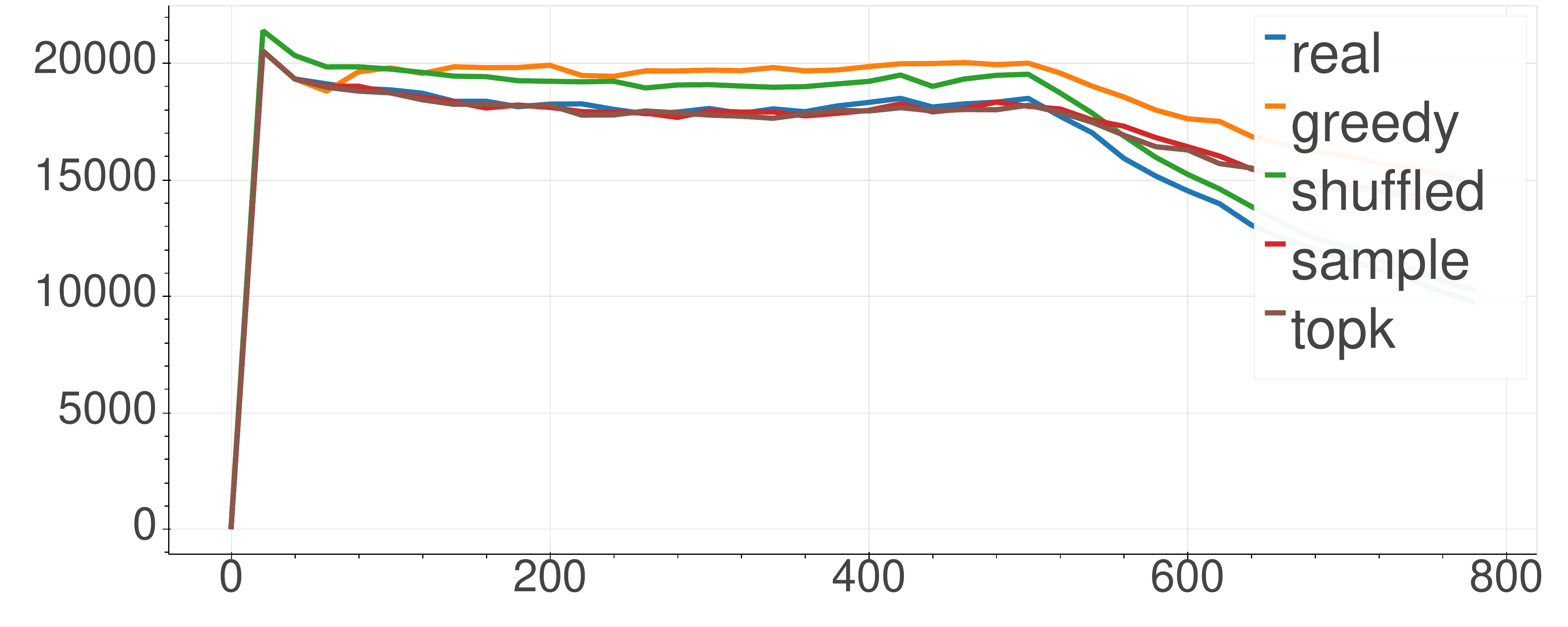}
    \includegraphics[width=0.49\linewidth]{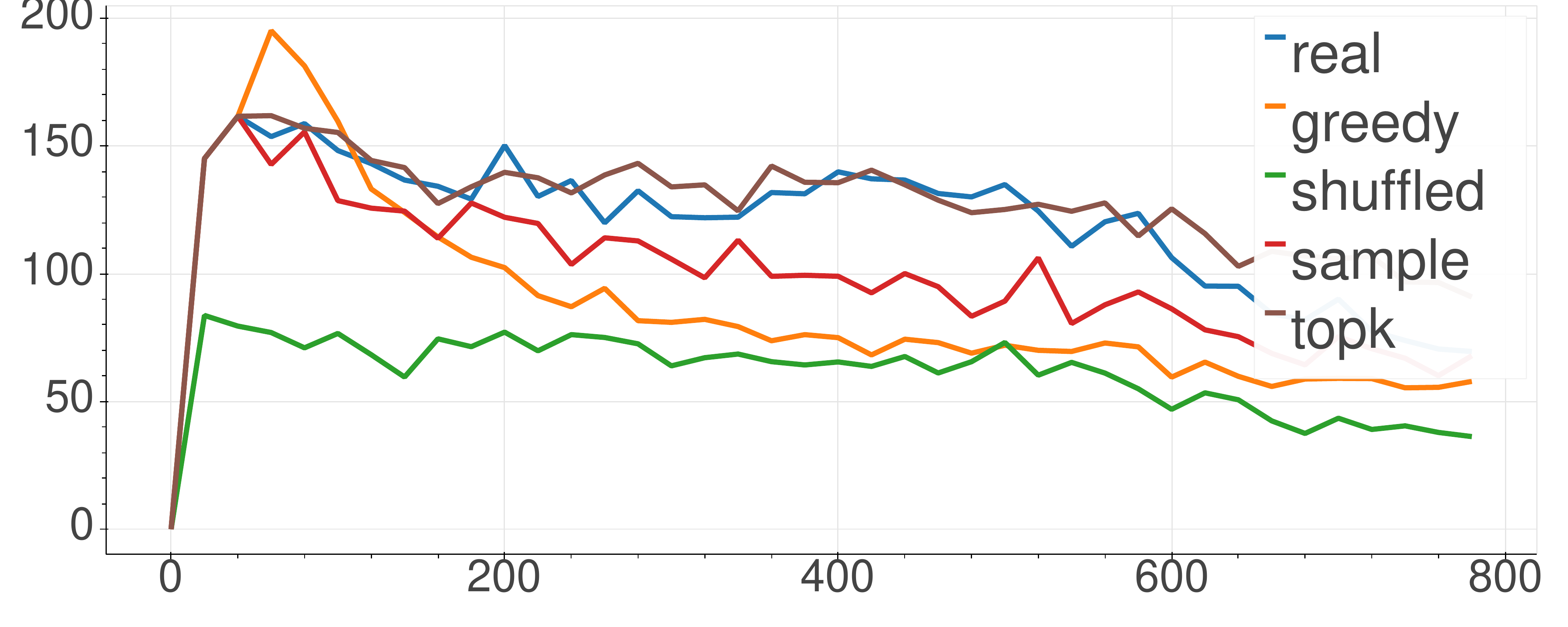}
    \includegraphics[width=0.49\linewidth]{figures/n-neighbor-valid-1024-layer-7.pdf}
    \includegraphics[width=0.49\linewidth]{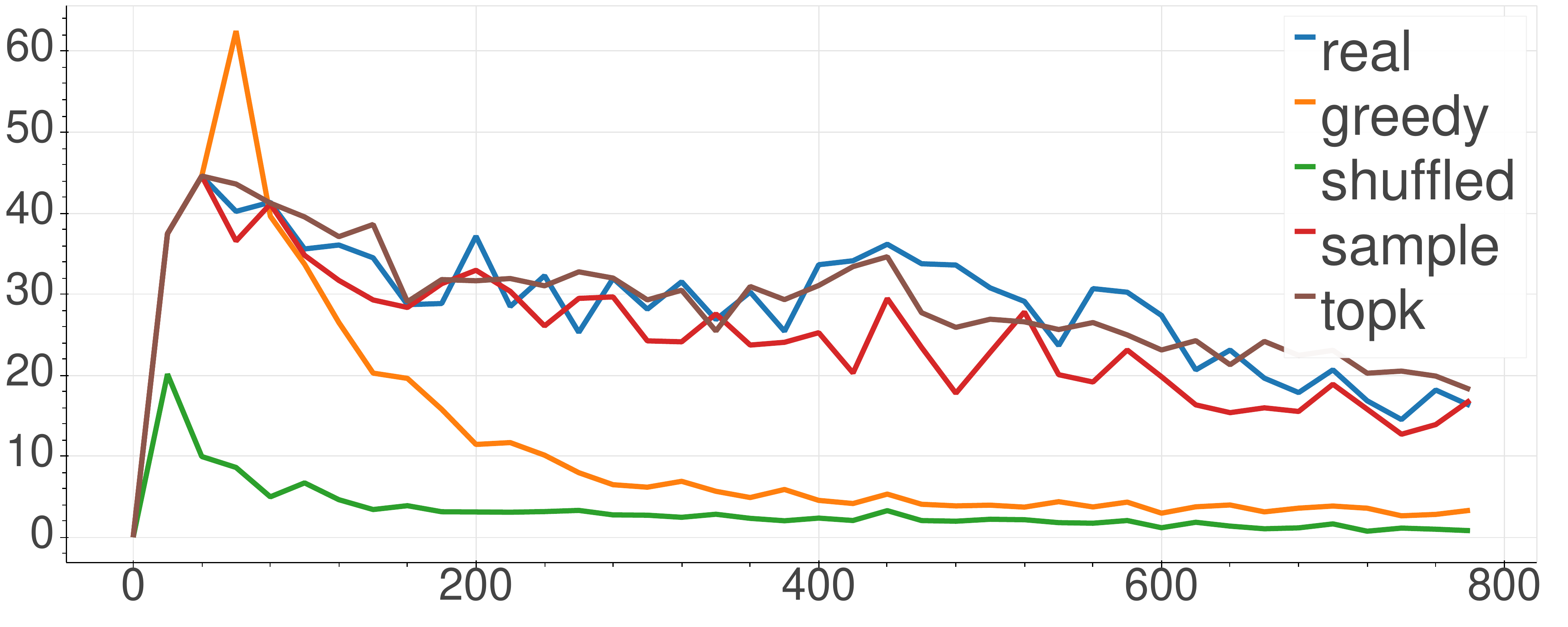}
    \includegraphics[width=0.49\linewidth]{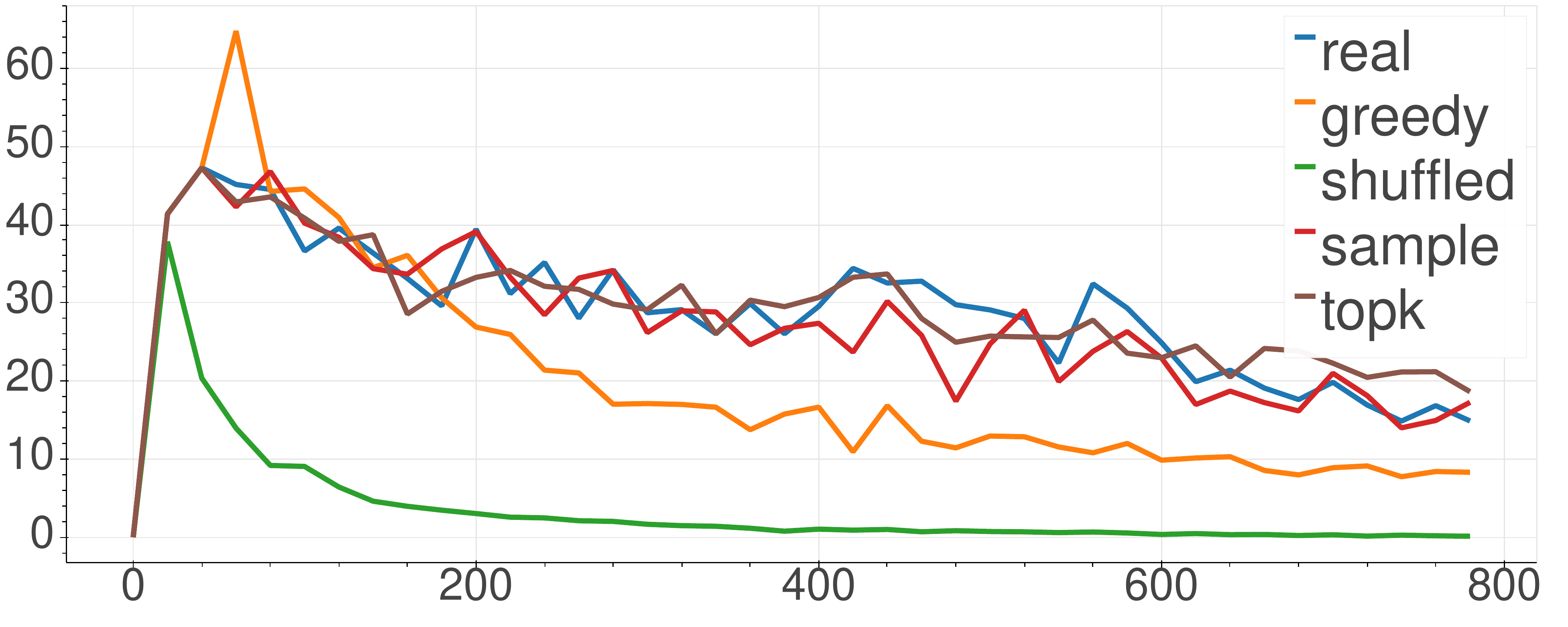}    \caption{Number of neighbors for \textit{compare-unseen} setting. The figures are the number of layers 1, 3, 5, 7, 9, 11, from left to right, top to bottom. The x-axis is the time step of the tokens. The y-axis is the number of real neighbors with the radius.}
    \label{fig:my_label}
\end{figure*}

\end{document}